\documentclass[sigconf]{acmart}
\renewcommand\footnotetextcopyrightpermission[1]{} 
\AtBeginDocument{%
  \providecommand\BibTeX{{%
    \normalfont B\kern-0.5em{\scshape i\kern-0.25em b}\kern-0.8em\TeX}}}

\usepackage{makecell, multirow, rotating}
\usepackage{subfigure}
\usepackage{graphicx}
\usepackage{textcomp,booktabs}
\usepackage{colortbl}
\usepackage{threeparttable}
\usepackage{tablefootnote}
\usepackage{fancyhdr}
\usepackage{balance}
\usepackage[]{hyperref}

\begin{document}
\fancyhead{}

\title{A Comprehensive Empirical Study of Vision-Language Pre-trained Model for Supervised Cross-Modal Retrieval }
\author{Zhixiong Zeng, Wenji Mao}
\author{Institute of Automation, Chinese Academy of Sciences}
\author{School of Artificial Intelligence, University of Chinese Academy of Sciences}
\email{{zengzhixiong2018, wenji.mao}@ia.ac.cn}
\begin{abstract} 
Cross-Modal Retrieval (CMR) is an important research topic across multimodal computing and information retrieval, which takes one type of data as the query to retrieve relevant data of another type. It has been widely used in many real-world applications. Recently, the vision-language pre-trained models represented by CLIP demonstrate its superiority in learning the visual and textual representations and gain impressive performance on various vision and language related tasks. Although CLIP as well as the previous pre-trained models have shown great performance improvement in the unsupervised CMR (\emph{i.e.}, cross-modal matching), the performance and impact of these pre-trained models on the supervised CMR were rarely explored due to the lack of common representation for the multimodal class-level associations. In this paper, we take CLIP as the current representative vision-language pre-trained model to conduct a comprehensive empirical study. We evaluate its performance and impact on the supervised CMR, and attempt to answer several key research questions. To this end, we first propose a novel model CLIP4CMR (CLIP enhanced network for Cross-Modal Retrieval) that employs the pre-trained CLIP as backbone network to perform the supervised CMR. Then by means of the CLIP4CMR framework, we revisit the design of different learning objectives in current CMR methods to provide new insights on model design. Moreover, we investigate the most concerned aspects in applying CMR, including the robustness to modality imbalance and sensitivity to hyper-parameters, to provide new perspectives for practical applications. Through extensive experiments, we show that CLIP4CMR achieves the SOTA results with prominent improvements on the benchmark datasets Wikipedia, NUS-WIDE, Pascal-Sentence and XmediaNet, and can be used as a fundamental framework to empirically study the key research issues of the supervised CMR, with significant implications for model design and practical considerations\footnote{Our data and codes are publicly available at \href{https://github.com/zhixiongz/CLIP4CMR}{https://github.com/zhixiongz/CLIP4CMR}.}.

\end{abstract}



%

\maketitle


\section{Introduction}
With the explosive increase of multimodal data in social media platforms, cross-modal retrieval (CMR) has become one of the emergent needs for people to acquire relevant images and texts conveniently. CMR is a fundamental task across multimodal computing and information retrieval, which takes the query in one modality to retrieve relevant data of another modality. It not only lays the basis for multimodal visual and language processing, analysis and understanding, but also facilitates a number of applications in domains such as image retrieval \cite{xia2014supervised}, image caption \cite{vinyals2015show}, recipe recommendation \cite{carvalho2018cross}, automatic story generation \cite{li2020topic} and so forth.

The aim of cross-modal retrieval is to establish the similarity link between samples from different modalities based on their semantic correlation. Existing research can be broadly categorized into two groups: the unsupervised CMR for paired multimodal data and the supervised CMR for labeled multimodal data. The unsupervised CMR (also called cross-modal matching) methods center on the design of explainable vision and language reasoning networks to learn the cross-modal semantic alignment, which gracefully aggregate the word-level and region-level fine-grained similarities into cross-modal similarity to perform the retrieval task \cite{chen2020imram,huang2017instance,song2019polysemous,wang2019camp,xu2015show,zhang2020context}. As the items from one modality usually have multiple semantic related items in another modality, the supervised CMR methods center on designing effective loss functions to preserve the multi-modal class-level semantic associations (\emph{i.e.}, the modality invariance and semantic discrimination) of the common representation space \cite{wang2017adversarial,wang2015,wu2018cycle,zeng2021pan,zhen2019deep,zheng2016hetero}. Due to the universality of multiple related samples across different modalities in reality, we focus on the supervised CMR in this paper, and use cross-modal matching and cross-modal retrieval to refer to the unsupervised and supervised CMR respectively.

Inspired by the great success of self-supervised pre-trained language models \cite{devlin2018bert,liu2019roberta}, a large number of vision-language pre-trained (VLP) models \cite{tan2019lxmert,lu2019vilbert,su2019vl,chen2019uniter,li2020oscar,desai2021virtex} have been developed that learn the vision-language semantic alignments to be finetuned on the downstream tasks. Recently, CLIP (Contrastive Language-Image Pre-training) \cite{radford2021learning} pre-trained on 400 million noisy multimodal web data has demonstrated its impressive performance on various downstream vision and language related tasks. The VLP models represented by CLIP are profoundly reshaping the cross-modal field \cite{cao2020behind} and their superiority on cross-modal tasks are increasingly recognized \cite{shin2021perspectives}. Although the VLP models have been successfully fine-tuned to the unsupervised cross-modal matching, their performance and impact on the supervised CMR have not been investigated, due to the fact that these pre-trained models cannot be directly applied to the supervised CMR, which requires the common representations of the more complex multimodal class-level associations.


In this paper, we conduct an empirical study of the vision-language pre-trained model for cross-modal retrieval. The first important research question raised for our empirical study is: can CLIP boost the performance of the CMR task and why? To explore this, we propose a model named CLIP4CMR, which takes the pre-trained CLIP as the backbone network. To generate the common representation space, CLIP4CMR exploits the pre-trained CLIP as the visual and textual encoders and then employs modality-specific multilayer perceptron for cross-modal retrieval. Although existing CMR methods rely heavily on the design of learning objectives, due to the diversity of model architectures, parameter choices and training protocols, previous research fails to supply a fair comparison vehicle for evaluating the learning objectives designed in the existing models. The CLIP4CMR framework provides a unified common ground for such fair comparison. The second important research question raised for our empirical study is: how does the design of different learning objectives (and their combination) influence the retrieval results? By means of CLIP4CMR, we are able to revisit the existing learning objectives, including the widely used pair-wise losses, more recent class-wise losses and hybrid ones that combine pair-wise and class-wise losses, and assess their comparative performances in the same experimental setting.

In addition, we consider the practical applications of the CMR models. Benefited from CLIP’s abundant multimodal knowledge obtained from extra pre-training data, we would like to investigate the practical aspects of applying the cross-modal retrieval model built on CLIP. The third important research question raised for our empirical study is: how does the CMR model built on CLIP perform under the practical situations? There are two key concerned issues here in practice: the robustness to modality imbalance \cite{zeng2021pan} and sensitivity to hyper-parameters \cite{luo2021clip4clip}, and therefore, the above research question is broken down into two sub-questions. The robustness to modality imbalance has attracted much attention recently due to the discrepancies of data collection and labor annotation between different modalities in real-world applications. To alleviate this problem, previous models are mainly based on the semantic consistency and modality heterogeneity to reconstruct modality-balanced data for improving robustness \cite{zeng2021pan,zeng2021mccn,jing2020incomplete}. The sensitivity to hyper-parameters is related to evaluating the scalability of a CMR model in real-world situations. In particular, the dimensionality of the common representation space is a crucial hyper-parameter for analyzing the computational storage and time efficiency of cross-modal retrieval, as usually the pre-calculated image and text representations are used for similarity ranking during the test phase. Previous studies have shown that the performance of the retrieval model in a more compact representation space is worse due to the lack of partial representation information \cite{roth2020revisiting,kim2021embedding}. With the new perspective brought in by CLIP, these issues need to be reexamined.

Through developing CLIP4CMR, this paper proposes the first supervised CMR framework built on the vision-language pre-trained model. Our empirical study based on CLIP4CMR contributes to cross-modal retrieval field in providing the following insights:
\begin{itemize}
	\item Benefited from the improvement of intra-class compactness, CLIP4CMR can significantly facilitate cross-modal retrieval task and serve as a promising new baseline.
	\item Under the unified experimental setting based on CLIP4CMR, currently widely-used hybrid losses that combine pair-wise and class-wise losses have no obvious performance gains compared to applying the class-wise loss alone.
	\item Cross-modal retrieval model built on CLIP can markedly improve the robustness to modality imbalance, and still maintain a small performance degradation in some extremely modality imbalanced cases. 
	\item Cross-modal retrieval model built on CLIP is almost insensitive to the dimension changes of the common representation space, and can still maintain relatively high performance in a very compact representation space.
\end{itemize}

\begin{figure*}\label{fig2}
	\centering
	\hspace*{-0.3cm}
	\includegraphics[width=1.03\linewidth]{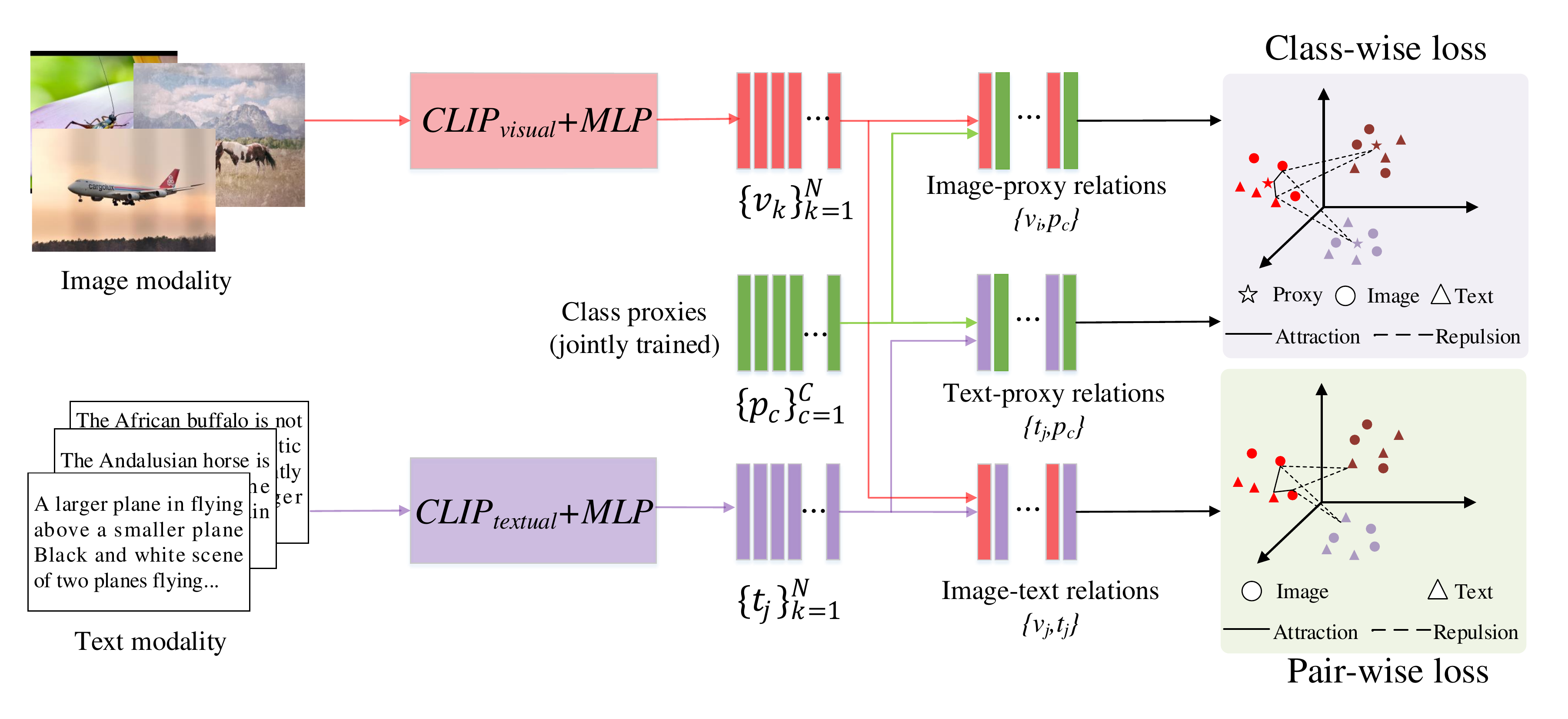}
	\caption{Overall architecture of the proposed CLIP4CMR. We leverage CLIP's visual encoder (\emph{i.e.}, CLIP$_{visual}$) and textual encoder (\emph{i.e.}, CLIP$_{textual}$) to generate original image and text representations and employ modality-specific multilayer perceptron layer (MLP) to learn common representation space. We then revisit the existing pair-wise and class-wise losses to provide insights on applying CLIP for supervised cross-modal retrieval.
	}
	\label{fig:2}
\end{figure*}

\section{Related Work}
Our work focuses on applying the vision-language pre-trained model to cross-modal retrieval task. Below we review cross-modal retrieval methods and vision-language pre-trained models.
\subsection{Cross-modal Retrieval}
The key challenge of cross-modal retrieval is to bridge the heterogeneity gap and learn transformation functions to project multimodal data into a common representation space, such that the cross-modal retrieval task boils down to the familiar nearest neighbor retrieval in the embedding space \cite{chun2021probabilistic}.
Existing cross-modal retrieval methods can be broadly categorized into two groups: the unsupervised methods for paired multimodal data and the supervised methods for labeled multimodal data.
The unsupervised methods focus on designing explainable vision and language reasoning networks to learn the cross-modal semantic alignment, which gracefully aggregate the word-level and region-level fine-grained similarities into cross-modal similarity to perform the retrieval task \cite{chen2020imram,huang2017instance,song2019polysemous,wang2019camp,xu2015show,zhang2020context}. 
The supervised methods focus on preserving the multimodal class-wise associations of the common representation space, so that the items of same class but from different modalities are closely grouped together \cite{wang2017adversarial,zeng2021pan,zhen2019deep,wang2016learning,wu2017joint,zeng2021mccn,wang2015,wu2020modality}.

The multimodal class-wise associations are mainly preserved by learning objectives for training the networks, including the widely used pair-wise losses, more recent class-wise losses and hybrid ones that combine pair-wise and class-wise losses.
The pair-wise loss provides rich class-level supervisory signals for learning
common representation space by comparing fine-grained intra-class and inter-class relations between items from different modalities, \emph{i.e.}, cross-modal data-to-data relations. A typical pair-wise loss is the modality invariant loss, which maximizes the intra-class similarities between items from different modalities \cite{hardoon2004canonical,feng2014cross,wang2015large,wang2015}.
Inspired by the success of deep metric learning in learning discriminative representations \cite{hu2014discriminative,bellet2013survey}, recent methods calculate the contrastive loss or semi-hard triplet loss on the multimodal data, thereby minimizing the similarity of intra-class multimodal pairs and maximizing that of inter-class multimodal pairs \cite{wang2017adversarial,zhen2019deep,peng2018modality,zeng2020event}.
In contrast, the class-wise loss leverages multimodal shared class proxies for learning common representation space by comparing samples with class proxies, \emph{i.e.}, data-to-proxy relations. 
The seminal examples are the linear regression loss \cite{wu2017joint,wang2015,zhen2019deep} and cross-entropy loss \cite{peng2016cross,wang2017adversarial,peng2019cm}, which project image and text samples into a shared label space to preserve class-level associations. 
Since the classification rule with softmax output layer lacks robustness to unknown classes \cite{yang2018robust}, the prototype contrastive loss has been proposed to improve the robustness issue by pulling samples towards the prototype of its class and pushing samples away from prototypes of other classes \cite{zeng2021mccn,zeng2021pan}.
In fact, most of the existing methods \cite{wang2017adversarial,wang2015,wu2017joint,zhen2019deep,peng2019cm} follow the paradigm of optimizing hybrid losses that combine pair-wise and class-wise losses to maximize information utilization, but fail to provide a fair comparison vehicle for evaluating the loss functions designed in these existing methods.

\subsection{Vision-Language Pre-trained Models}
Recently, self-supervised language pre-trained models such as BERT \cite{devlin2018bert}, RoBERTa \cite{liu2019roberta}, GPT2 \cite{radford2019language} have pushed the state of the art on a wide range of NLP tasks. There are two keys to their success: effective pre-training tasks over large-scale language corpus,
and the utilization of Transformer \cite{vaswani2017attention} for learning contextualized text representations \cite{chen2019uniter}.
Inspired by the success of pre-trained language models, a large number of vision-language pre-trained (VLP) models \cite{tan2019lxmert,lu2019vilbert,su2019vl,chen2019uniter,li2020oscar,desai2021virtex,geigle2021retrieve,sun2021lightningdot,radford2021learning} based on Transformer have been made to build the multimodal counterpart that learns vision-language semantic alignments, bringing about great advances on downstream multimodal tasks like cross-modal retrieval. 

Exemplary VLP models can be categorized as cross-encoder based and embedding based methods \cite{geigle2021retrieve}.
The cross-encoder based methods \cite{tan2019lxmert,lu2019vilbert,su2019vl,chen2019uniter,li2020oscar} apply
a cross-attention mechanism based on Transformer-based neural architectures to compute the similarity score between items from different modalities. The embedding based methods encode multimodal items separately to generate high-dimensional visual and textual representations, and utilize the standard distance metrics to compute the cross-modal similarity \cite{desai2021virtex,geigle2021retrieve,sun2021lightningdot,radford2021learning}. More recently, the CLIP \cite{radford2021learning} employs the embedding-based architecture and is pre-trained on $400$ million noisy multimodal web data, and achieves impressive performance on many downstream vision and language related tasks. The great success of CLIP comes from the generality and usability learned from hundreds of millions of raw image and text data. It has inspired the growing interest of empirical studies that explore the impact of CLIP on video retrieval \cite{luo2021clip4clip}, visual question answering and visual entailment \cite{shen2021much}.

\section{The Unified Framework}
Figure 1 illustrates the unified framework of applying the vision-language pre-trained model for cross-modal retrieval, which consists of the design of CLIP4CMR model and learning objectives. 

\subsection{Design of CLIP4CMR}\label{sec:3.1}
Without losing generality, we focus on cross-modal retrieval for image and text. 
Suppose that we have a collection of $N$ instances of image-text pairs, denoted as $\Psi_P$ = $\{(x_k^v, x_k^t)\}_{k=1}^{N}$, where $x_k^v$ is the input image sample and $x_k^t$ is the input text sample. Each pair $(x_k^v, x_k^t)$ is assigned a semantic label $y_k \in \{1, 2, ..., C\} $, where $C$ is the number of semantic categories. 

Inspired by the superiority of CLIP in learning visual and textual representations, we utilize the model architecture of CLIP to perform cross-modal retrieval. The model architecture of CLIP consists of a visual encoder for image modality and a textual encoder for text modality. 
The visual encoder takes the form of the convolutional neural network like ResNet-50 \cite{he2016deep} or vision transformers like ViT \cite{dosovitskiy2020image}, and is pre-trained by a broad source of textual supervision to learn low-dimensional image representations.
 The textual encoder is built on top of a Transformer \cite{vaswani2017attention}, and is pre-trained by a broad source of visual supervision to learn low-dimensional text representations. 
 We employ the pre-trained CLIP to generate image and text representations, which can be formulated as:
\begin{equation}
{\alpha_i} = CLIP_{v}(x_i^v), \quad {\beta_j} = CLIP_{t}(x_j^t)
\end{equation}
where both $\alpha_i$ and $\beta_j$ are $1024$-dimensional representations, and CLIP$_{v}$ and CLIP$_{t}$ denote the visual encoder and textual encoder of CLIP, respectively. 
However, it may be unreasonable to directly apply the representation space generated by CLIP for cross-modal retrieval, as CLIP pre-trained by self-supervised task fails to capture the more complex class-level semantic discrimination.
Thus we deploy modality-specific multilayer perceptron to generate a common representation space as in most existing work \cite{zeng2021mccn,zeng2021pan}, which can be formulated as:
\begin{equation}
{v_i} = W_{v2}(\sigma(W_{v1}\alpha_i+b_{v1})) + b_{v2}
\end{equation}
\begin{equation}
{t_j} = W_{t2}(\sigma(W_{t1}\beta_i+b_{t1})) + b_{t2}
\end{equation}
where $\sigma$  denotes the GeLU \cite{hendrycks2016gaussian} activation function, $W_{v1}$, $W_{v2}$, $b_{v1}$, $b_{v2}$, $W_{t1}$, $W_{t2}$, $b_{t1}$ and $b_{t2}$ are the trainable parameters, ${v_i} \in\mathbb{R}^{d}$ and ${t_j} \in\mathbb{R}^{d}$ are the projected features in the common representation space, and $d$ is the dimension of the representation space. To prevent the divergence of the magnitudes, we apply l2-normalization layer to output the normalized representations.

\subsection{Learning Objectives}\label{sec:3.1}
\subsubsection{Pair-wise loss}
The pair-wise loss provides rich supervisory signals for learning
common representation space by comparing fine-grained intra-class and inter-class relations between samples from different modalities, \emph{i.e.}, cross-modal data-to-data relations.
A seminal pair-wise loss for cross-modal retrieval is the contrastive loss, which minimizes the distances of positive image-text pairs belonging to the same class and maximizes the distances of negative pairs for being larger than a margin \cite{wang2015,wu2017joint,peng2017ccl}. Given a batch of $n$ image-text pairs, it can be formulated as:
\begin{equation}
\mathcal{L}_{cl} = \frac{1}{n}\sum_{i=1}^{n}\sum_{j=1}^{n}y_{i,j}d(v_i,t_j)+\frac{1}{n}\sum_{i=1}^{n}\sum_{j=1}^{n}(1-y_{i,j})[\Delta-d(v_i,t_j)]_+
\end{equation}
where $d(\cdot,\cdot)$ denotes the square of the Euclidean distance, $\Delta$ denotes the distance margin, and the label $y_{i,j} \in \{0,1\}$ indicates whether an image-text pair $(v_i,t_j)$ belongs to the same class or not.
Some early cross-modal retrieval methods \cite{feng2014cross,zhai2013learning} only consider the optimization of positive image-text pairs in Equation (4), which was called modality-invariant loss in subsequent work \cite{zhen2019deep}. Another popular pair-loss for cross-modal retrieval is the triplet loss, which encourages the distances of positive image-text pairs to be smaller than that of negative pairs with a margin $\Delta$ \cite{peng2018modality,wang2017adversarial}.
Given a batch of $n$ image-text pairs, it can be formulated as:
\begin{equation}
\begin{split}
\mathcal{L}_{tl} &= \frac{1}{|X_v|}\sum_{(v_i,t_j^+,t_k^-) \in X_v}[d(v_i,t_j^+)-d(v_i,t_k^-)+\Delta]_+ \\ &+ \frac{1}{|X_t|}\sum_{(t_i,v_j^+,v_k^-) \in X_t}[d(t_i,v_j^+)-d(t_i,v_k^-)+\Delta]_+
\end{split}
\end{equation}
where $X_v$ denotes the set of triplets by select $v_i$ as anchor to find positive text $t_j^+$ and negative text $t_k^-$, $X_t$ denotes the set of triplets by select $t_i$ as anchor to find positive image $v_j^+$ and negative image $v_k^-$, $|X_v|$ and $|X_t|$ are their cardinalities.

\begin{table*}[tbp]
	\renewcommand\arraystretch{1.2}   
	\centering
	\caption{Performance comparison in terms of mAP on four widely-used benchmark datasets for cross-modal retrieval.}
	\setlength{\tabcolsep}{2.25mm}{
		\begin{tabular}{ccccccccccccc}
			\hline
			\multirow{2}[4]{*}{MAP} & \multicolumn{3}{c}{Wikipedia} & \multicolumn{3}{c}{Pascal-Sentence} & \multicolumn{3}{c}{NUS-WIDE} & \multicolumn{3}{c}{XmediaNet} \\
			\cmidrule(r){2-4} 	\cmidrule(r){5-7} \cmidrule(r){8-10} \cmidrule(r){11-13}   & I2T & T2I & Avg. & I2T & T2I & Avg. & I2T & T2I & Avg. & I2T & T2I & Avg.  \\
			\hline
			CCA \cite{hardoon2004canonical} &0.298&0.273&0.286 & 0.203 & 0.208  & 0.206  & 0.167  & 0.181  & 0.174  & 0.212 & 0.217 & 0.215 \\
			KCCA \cite{wang2015large}&0.438&0.389&0.414 & 0.488  & 0.446  & 0.467  & 0.351  & 0.356  & 0.354  & 0.252 & 0.27  & 0.261 \\
			
			Corr-AE \cite{feng2014cross}&0.442&0.429&0.436 & 0.532  & 0.521  & 0.527  & 0.441  & 0.494  & 0.468  & 0.469 & 0.507 & 0.488 \\
			
			JRL \cite{zhai2013learning}  &0.479&0.428&0.454 & 0.563  & 0.505  & 0.534  & 0.466  & 0.499  & 0.483  & 0.488 & 0.405 & 0.447 \\
			CMDN \cite{peng2016cross} &0.487&0.427&0.457 & 0.544  & 0.526  & 0.535  & 0.492  & 0.542  & 0.517  & 0.485 & 0.516 & 0.501 \\
			JFSSL \cite{wang2015}&0.458&0.426&0.442 & 0.553  & 0.542  & 0.548  & 0.514  & 0.523  & 0.519  & 0.525 & 0.518 & 0.521 \\
			ACMR \cite{wang2017adversarial}  &0.468&0.412&0.440& 0.538  & 0.544  & 0.541  & 0.519  & 0.542  & 0.531  & 0.536 & 0.519 & 0.528 \\
			JLSLR \cite{wu2017joint}&0.473&0.440&0.456 & 0.568  & 0.551  & 0.560  & 0.536  & 0.531  & 0.534  & 0.544 & 0.553 & 0.549 \\
			MCSM \cite{peng2018modality}&0.516&0.458&0.487 & 0.598  & 0.598  & 0.598  & 0.522  &  0.546 & 0.534 & 0.540  & 0.550  & 0.545 \\
			CCL \cite{peng2017ccl}&0.505&0.457&0.481 & 0.576  & 0.561  & 0.569  & 0.506  &  0.535 & 0.521 & 0.537  & 0.528  & 0.533 \\
			CM-GANS \cite{peng2019cm} &0.521&0.466&0.494& 0.603  & 0.604  & 0.604  &  0.536 &  0.551  & 0.543  & 0.567 & 0.551 & 0.559 \\
			PAN \cite{zeng2021pan} &0.517&0.462&0.489& 0.686  & 0.689  & 0.688  & 0.590  & 0.571  & 0.581  & 0.669 & 0.660  & 0.665 \\
			DSCMR$^*$ \cite{zhen2019deep}&0.521&0.478&0.499& 0.674  & 0.682 & 0.678$^\dagger$ & 0.611  & 0.615   & 0.613  & 0.697 & 0.693 & 0.695 \\
			MCCN$^*$ \cite{zeng2021mccn} &0.552&0.487&0.520& 0.681  & 0.686  & 0.683$^\dagger$  &-  & -  & -  & 0.741& 0.743  & 0.742 \\
			\hline \textbf{CLIP4CMR} & \textbf{0.592} & \textbf{0.574}  & \textbf{0.583}  & \textbf{0.698}  & \textbf{0.692}  & \textbf{0.695}  & \textbf{0.609}  & \textbf{0.621}  & \textbf{0.615 } &  \textbf{  0.746}   &     \textbf{0.758}  & \textbf{0.752} \\		
			\hline
	\end{tabular}}%
		\begin{tablenotes}
			\footnotesize
			\item $^*$ Two-stage approach, which use training data to train pre-classified visual and textual encoders followed by cross-modal retrieval.
			\item $^\dagger$ Reproducible results using $200$ test samples in Pascal-Sentence dataset following \cite{wang2017adversarial}. The average mAP of our method is $74.2$ when following the dataset split of MCCN \cite{zeng2021mccn}, but their test samples are too small for effective evaluation.
		\end{tablenotes}
	\label{tab:addlabel}%
\end{table*}%

\subsubsection{Class-wise loss}
The class-wise loss leverage multimodal shared class proxies for learning common representation space by comparing samples with class proxies, \emph{i.e.}, data-to-proxy relations. 
A seminal example is the linear regression loss, which can be formulated as \cite{wu2017joint,zhen2019deep,wang2015}:
\begin{equation}
\mathcal{L}_{lrl} = \dfrac{1}{\emph{n}}\sum_{i}(\Vert{Q^Tv_i - Y_i}\Vert_2 + \Vert{Q^Tt_i - Y_i}\Vert_2)
\end{equation}
where $\Vert{.}\Vert_F$ denotes the Frobenius norm, $Q$ is the projection matrix of the linear classifier, $Y_i$ is the one-hot label vector where the $y_i$-th element is 1 and the others are 0. Each column of the projection matrix $Q$ represents a class proxy, which provides a unified anchor to pull together all images and texts belonging to the same class.
To exploit the nonlinearity of the label space, another popular class-wise loss is the cross-entropy loss calculated by \cite{wang2017adversarial,peng2019cm}:
\begin{equation}
\mathcal{L}_{cel} = -\frac{1}{n}\sum_{i=1}^{n}[log\frac{e^{W^T_{y_i}v_i+b_{y_i}}}{\sum_{j=1}^{C}e^{W^T_jv_i+b_j}}+log\frac{e^{W^T_{y_i}t_i+b_{y_i}}}{\sum_{j=1}^{C}e^{W^T_jt_i+b_j}}]
\end{equation}
where $W_j$ and $b_j$ denote the $j$-th column of the weight matrix and bias matrix of the shared classification layer. Here the layer parameters $W_j$ and $b_j$ can be regarded as a class proxy with bias term. However, this classification rule with softmax output layer lacks robustness to unknown classes \cite{yang2018robust}. To improve the robustness of cross-modal retrieval, recent work PAN \cite{zeng2021pan} assigns a set of unified prototypes $P=\{p_c|c=1,2,...,C\}$ as class proxies and adopt the nearest-prototype classification rule to infer unknown classes. The multimodal representations and prototypes are jointly learned through a prototype contrastive loss:
\begin{equation}\label{eq7}
\mathcal{L}_{pcl} = -\frac{1}{n}\sum_{i=1}^{n}[log\frac{e^{-\lambda d(v_i-p_{y_i})}}{\sum_{j=1}^{C}e^{-\lambda d(v_i-p_{j})}}+log\frac{e^{-\lambda d(t_i-p_{y_i})}}{\sum_{j=1}^{C}e^{-\lambda d(t_i-p_{j})}}]
\end{equation}
here $\lambda$ is a scaling factor.


\subsubsection{hybrid loss}
To utilize both data-to-data and data-to-proxy relations and maximize information utilization, most of the existing methods \cite{wang2017adversarial,wang2015,wu2017joint,zhen2019deep,peng2019cm} follow the paradigm of optimizing hybrid losses that combine class-wise and pair-wise losses.
Generally, the hybrid loss can be formulated as: 
\begin{equation}
\mathcal{L}_{hybrid}=\mathcal{L}_{class-wise}+\gamma \mathcal{L}_{pair-wise} 
\end{equation}
where $\mathcal{L}_{class-wise} \in \{\mathcal{L}_{lrl}, \mathcal{L}_{cel}, \mathcal{L}_{pcl}\}$,  $\mathcal{L}_{pair-wise} \in \{\mathcal{L}_{cl}, \mathcal{L}_{tl}\}$, and $\gamma$ is a carefully selected combination weight.

\begin{figure*}[h]\label{fig5}
	\begin{flushleft}		
		\centering 
		\subfigure[Wikipedia intra-class distances]{ 
			\raggedright{\includegraphics[width=0.241\textwidth]{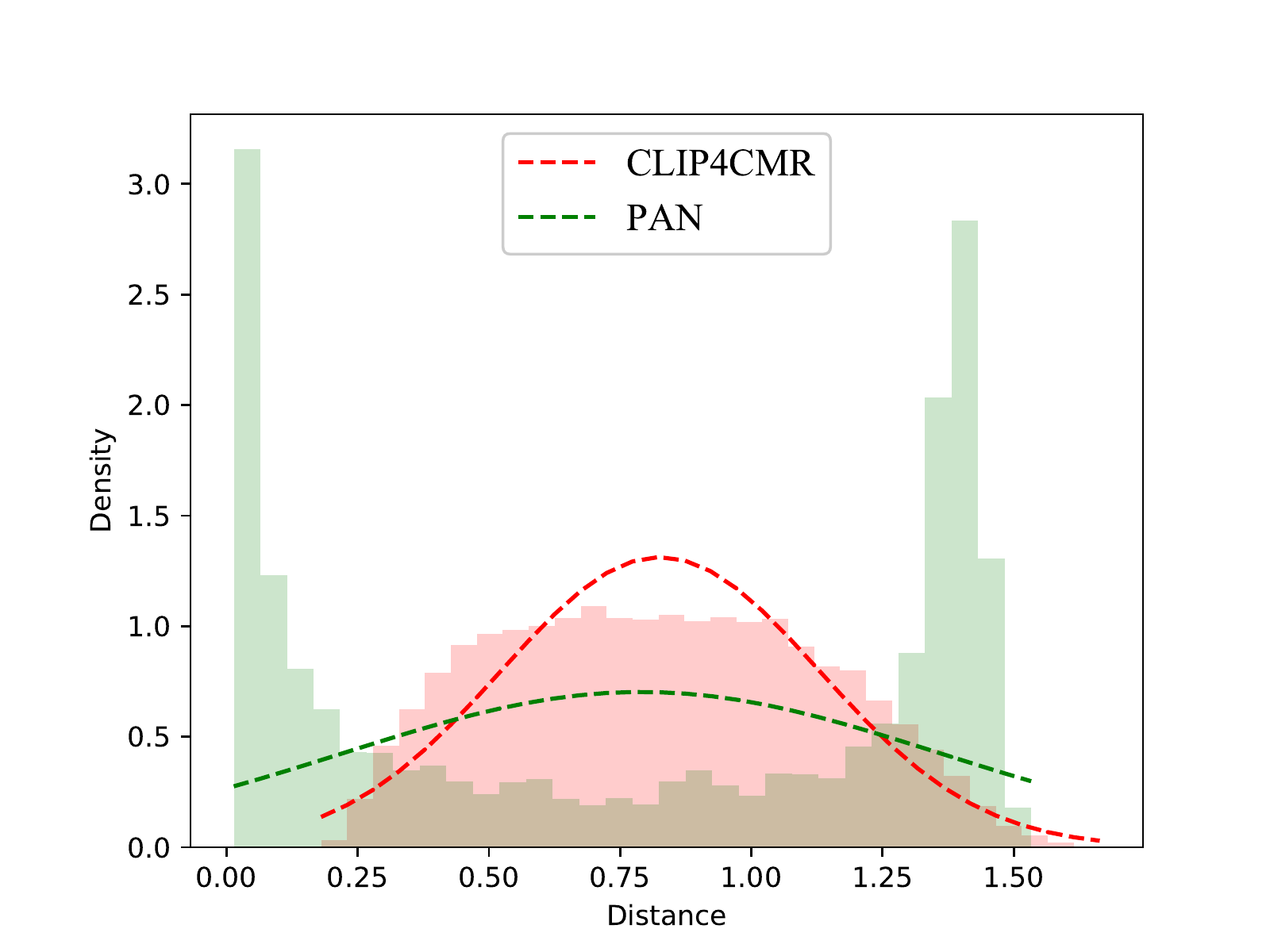}}}
		\subfigure[Wikipedia inter-class distances]{ 
			\raggedright{\includegraphics[width=0.241\textwidth]{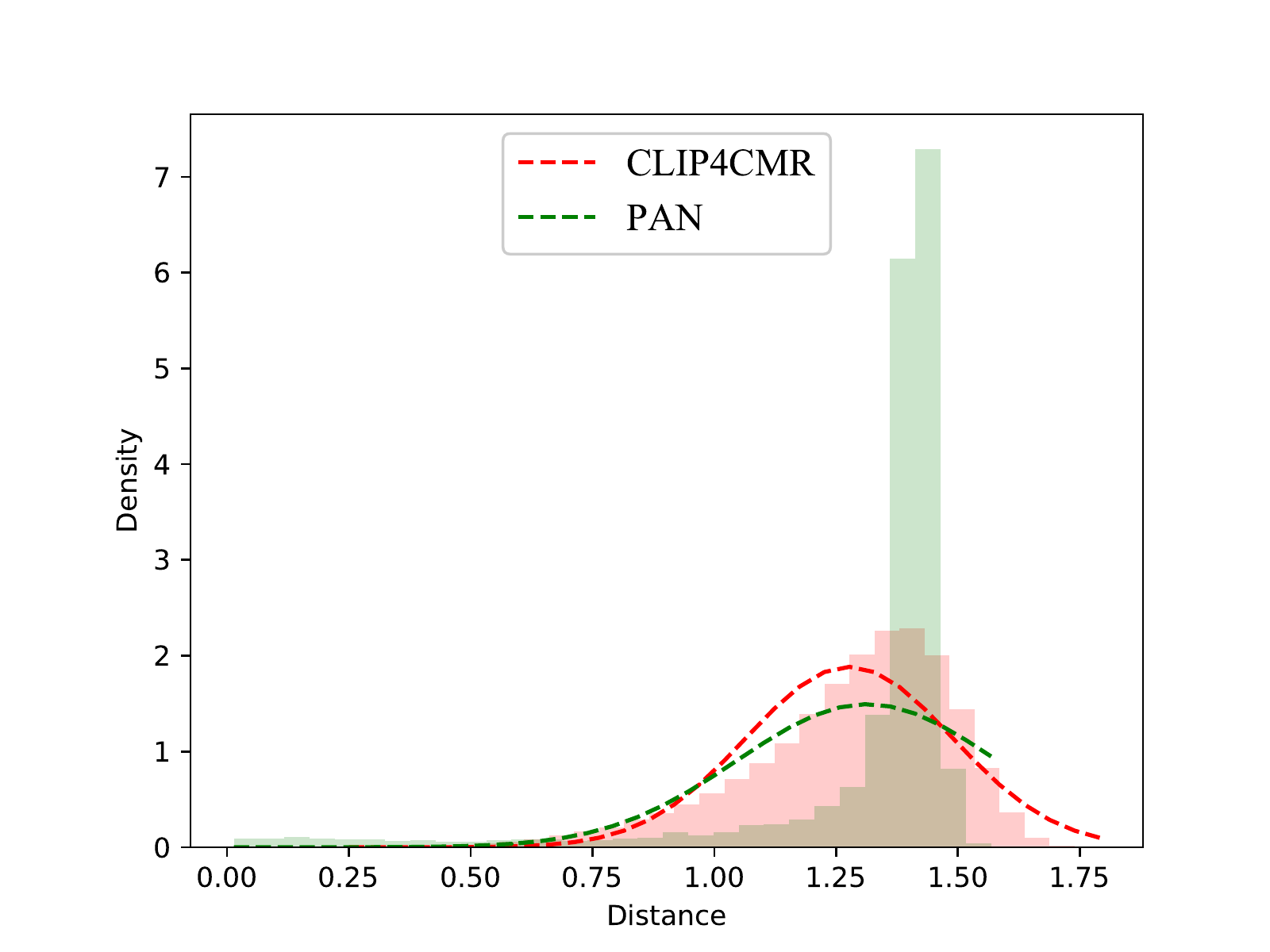}}}
		\subfigure[XmediaNet intra-class distances]{ 
			\raggedright{\includegraphics[width=0.241\textwidth]{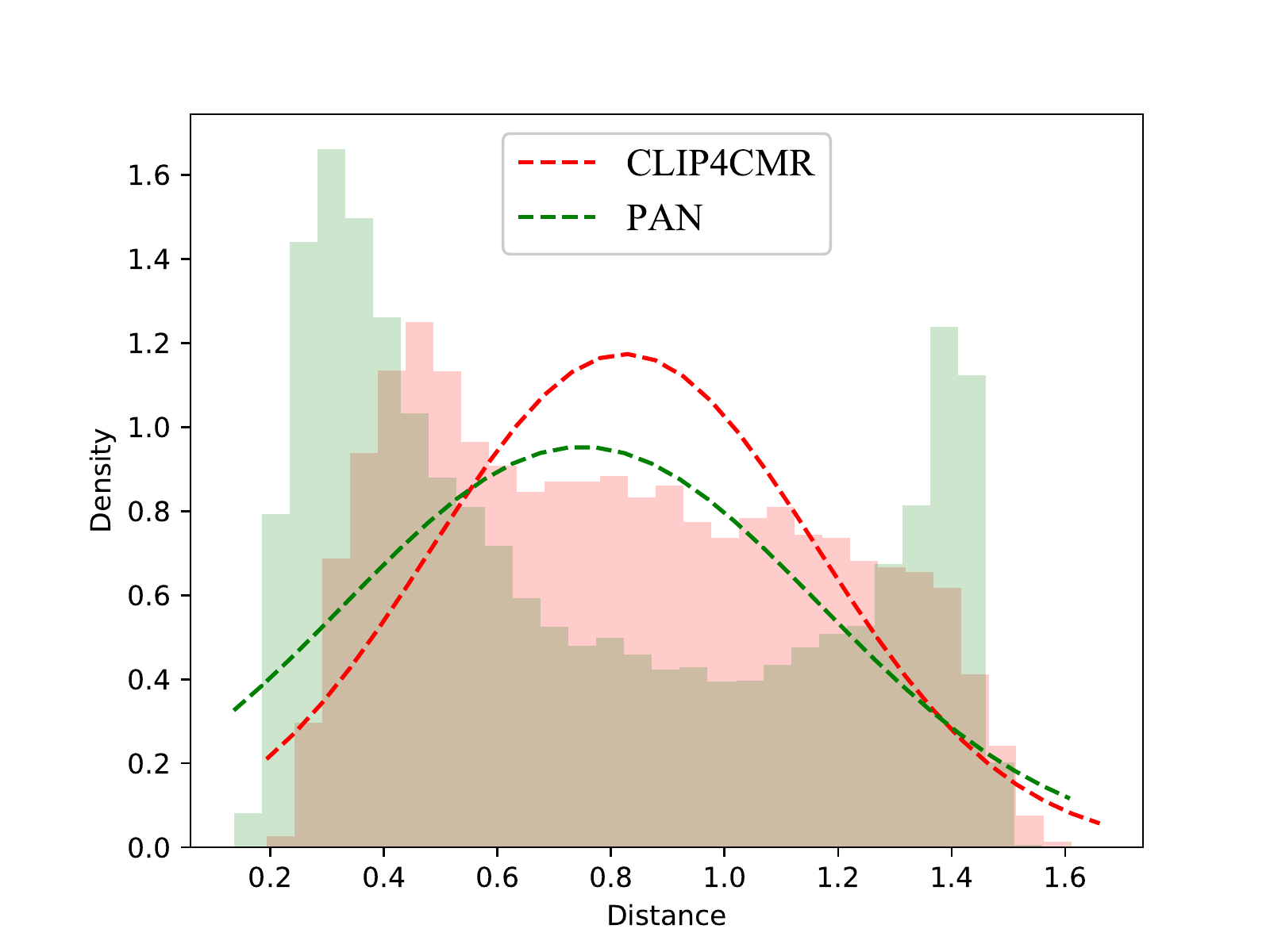}}}
		\subfigure[XmediaNet inter-class distances]{ 
			\raggedright{\includegraphics[width=0.241\textwidth]{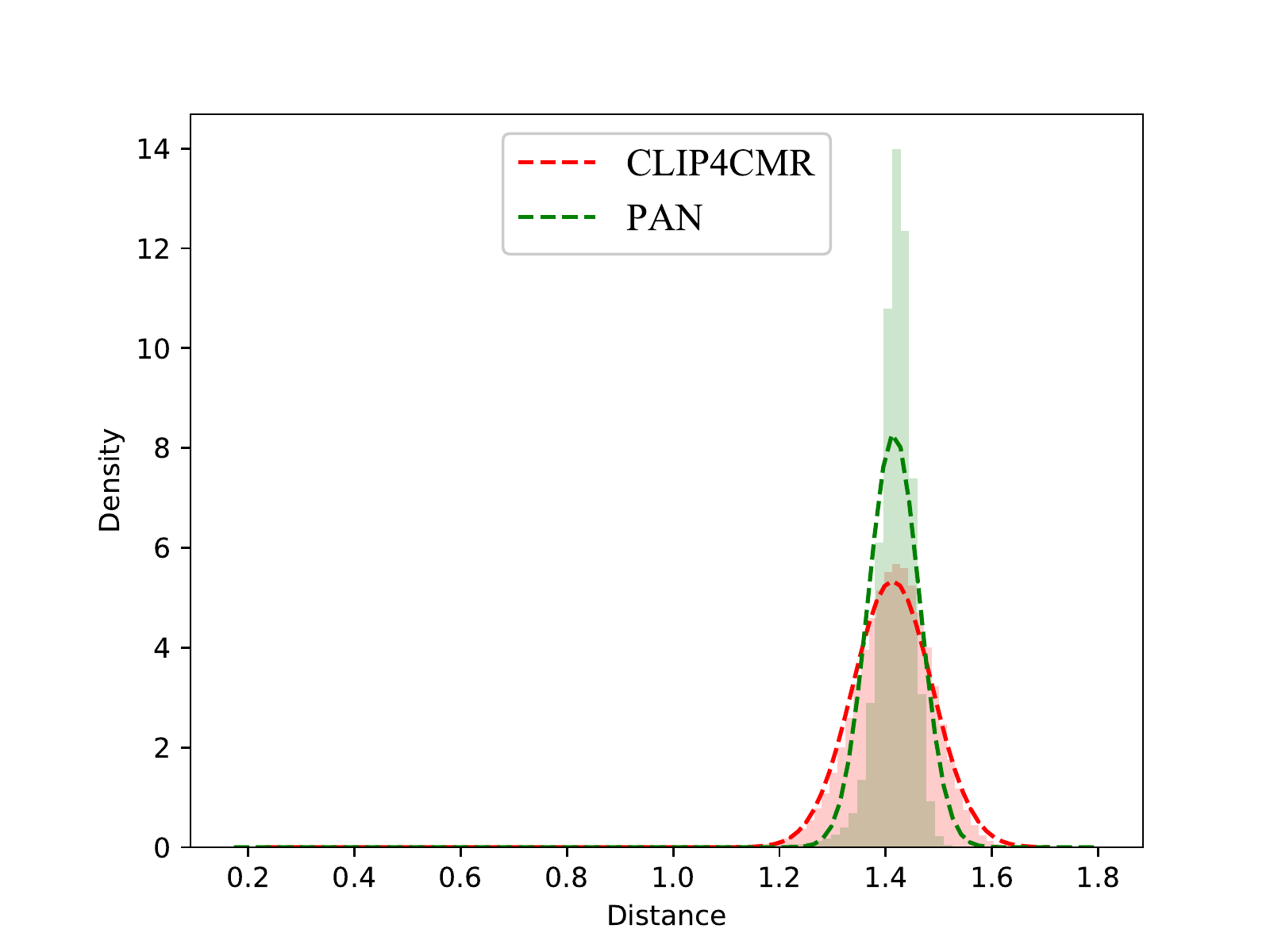}}}
		\caption{The distributions of the intra-class distances and inter-class distances across different modalities in the test set.} 
		\label{multihead} 
	\end{flushleft}
\end{figure*}

\section{Experiments}
\subsection{Experimental Setup} \label{sec:4.1}
\subsubsection{Datasets}

To verify the effectiveness of our proposed method, we conduct our empirical study on four widely-used benchmark datasets, namely Wikipedia \cite{rasiwasia2010new}, Pascal-Sentence \cite{rashtchian2010collecting}, NUS-WIDE\cite{chua2009nus} and XmediaNet \cite{peng2018modality}. For the Wikipedia dataset, we use 2,157 image-text pairs from 10 semantic classes for training, and 462 image-text pairs for test.
For the Pascal-Sentence dataset, we use 8,00 image-text pairs from 20 classes for training and 200 image-text pairs for test. For the NUS-WIDE dataset, we use 8,000 image-text pairs from 10 classes for training and 1,000 image-text pairs for test. For the XmediaNet dataset, we use 32,000 image-text pairs from 200 classes for training and the other 4,000 image-text pairs for test. The dataset splits mainly follow those in \cite{zhen2019deep,wang2017adversarial}.

\begin{table*}[tbp]
	\renewcommand\arraystretch{1.35}   
	\centering
	\caption{Revisiting pair-wise and class-wise losses in cross-modal retrieval with the unified CLIP4CMR framework.}
	\setlength{\tabcolsep}{2.30mm}{
		\begin{tabular}{cccccccccccccc}
			\hline
			\multicolumn{2}{c}{\multirow{2}[4]{*}{MAP}} & \multicolumn{3}{c}{Wikipedia} & \multicolumn{3}{c}{Pascal-Sentence} & \multicolumn{3}{c}{NUS-WIDE} & \multicolumn{3}{c}{XmediaNet} \\
			\cmidrule(r){3-5} 	\cmidrule(r){6-8} \cmidrule(r){9-11} \cmidrule(r){12-14}     \multicolumn{2}{c}{} & I2T & T2I & Avg. & I2T & T2I & Avg. & I2T & T2I & Avg. & I2T & T2I & Avg. \\
			\hline
			
			\multirow{3}[2]{*}{Class-wise loss} 
			& LRL & \textbf{0.592}  &\textbf{ 0.585}  & \textbf{0.588}  & 0.686  & 0.680  & 0.683  & \textbf{0.621}  & \textbf{0.643}  & \textbf{0.632}  & 0.574  & 0.576  & 0.575  \\
			
			& CEL & 0.586  & 0.565  & 0.576  & 0.697  & 0.686& 0.692  & 0.605  & 0.619 &  0.612 & 0.671 & 0.674 & 0.673 \\  
			& PCL & \textbf{0.592} & 0.574  & 0.583  & \textbf{0.698}  & \textbf{0.692} & \textbf{0.695}  & 0.609  & 0.621  & 0.615 &   \textbf{0.746}   & \textbf{0.758} & \textbf{0.752} \\
			\hline
			\multirow{3}[2]{*}{Pair-wise loss} 
			& ML & 0.147  & 0.153  & 0.150  & 0.114  & 0.104  & 0.109  & 0.137  &  0.131&  0.134 & 0.012  &  0.011 & 0.012 \\
			& CL & 0.516  & 0.498  & 0.507  & 0.587  & 0.555  & 0.571  & 0.577  &  0.592&  0.584 & 0.628  & 0.641  & 0.635  \\
			& TL & 0.550 &  0.536 & 0.543  &0.624  &0.620   & 0.622 &  0.595 & 0.603 &  0.599 &  0.674 & 0.678  & 0.676 \\
			
			\hline
	\end{tabular}}%
	\label{tab:addlabel}%
\end{table*}%
\begin{figure*}[tbp]\label{fig5}
	\begin{flushleft}		
		\centering 
		\subfigure[Wikipedia, LRL]{ 
			\raggedright{\includegraphics[width=0.261\textwidth]{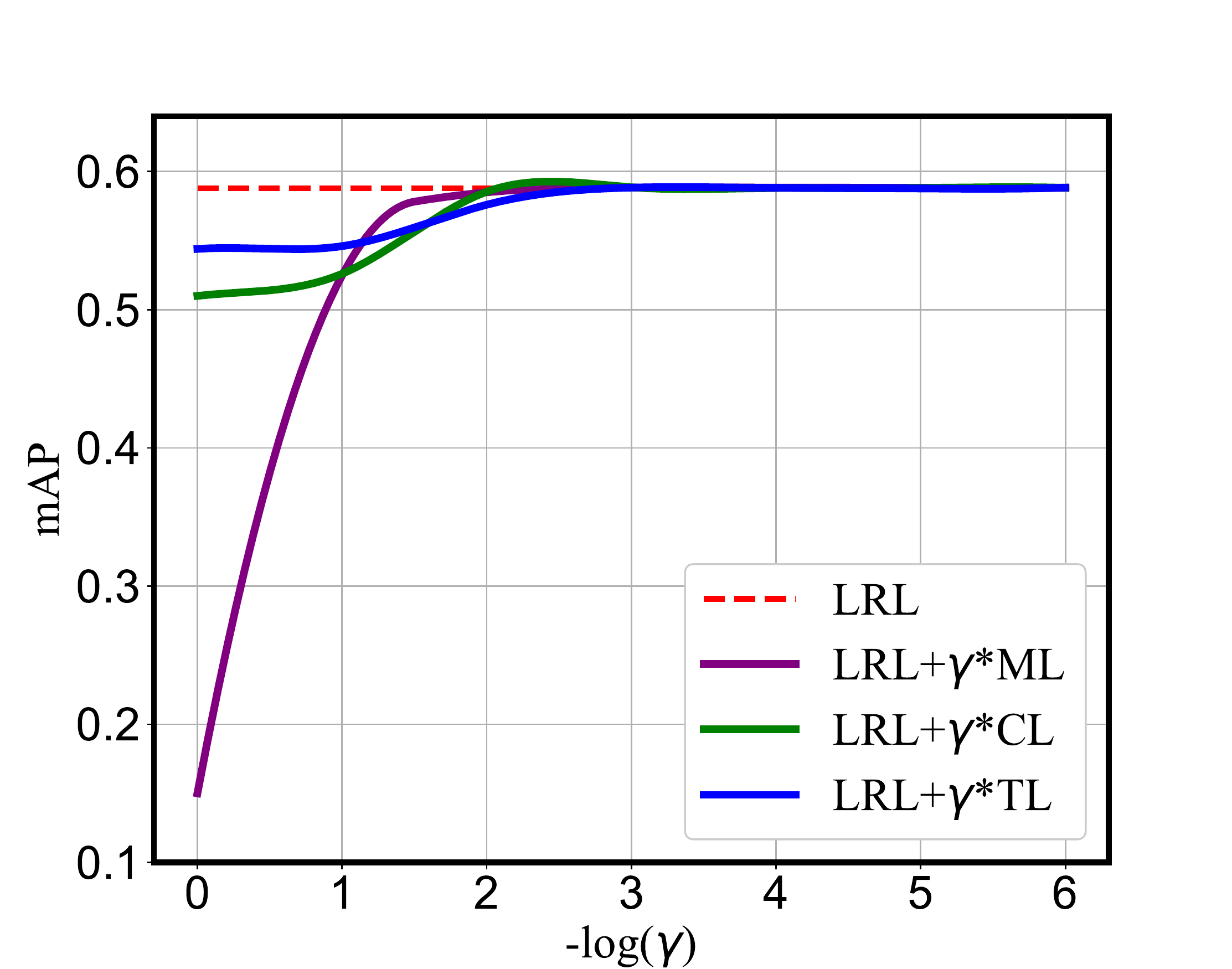}}}
		\subfigure[Wikipedia, CEL]{ 
			\raggedright{\includegraphics[width=0.261\textwidth]{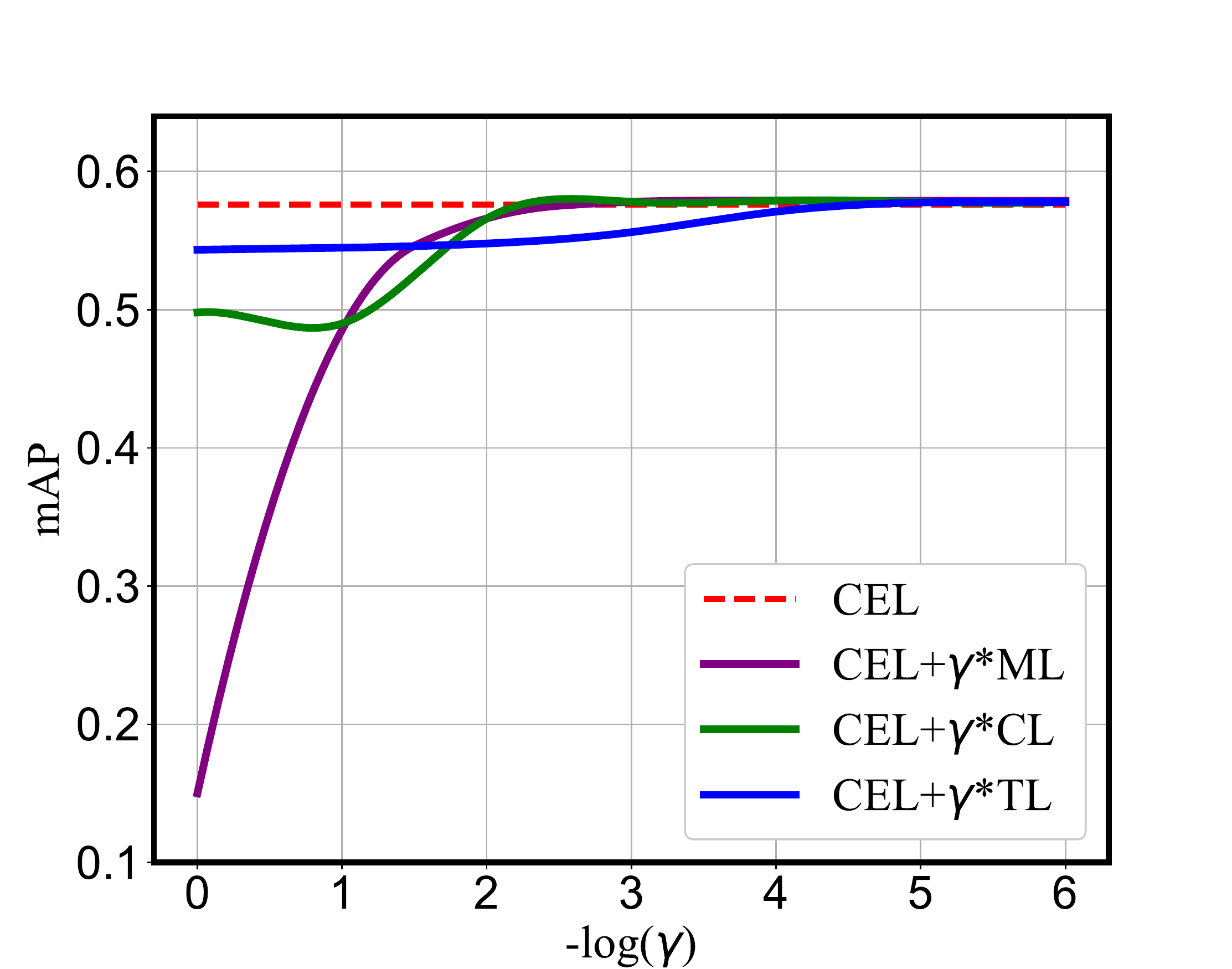}}}
		\subfigure[Wikipedia, PCL]{ 
			\raggedright{\includegraphics[width=0.261\textwidth]{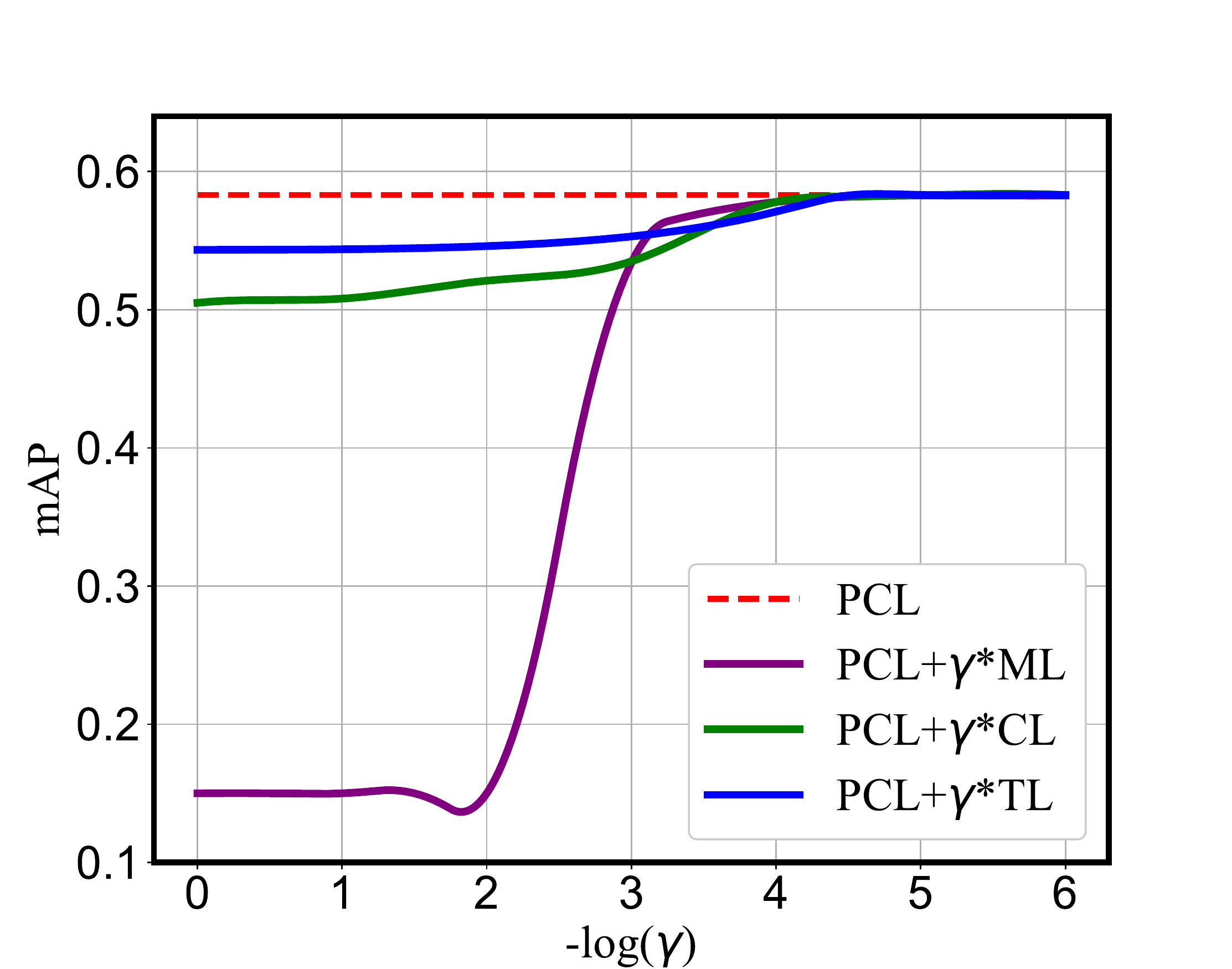}}}
		\subfigure[NUS-WIDE, LRL]{ 
			\raggedright{\includegraphics[width=0.261\textwidth]{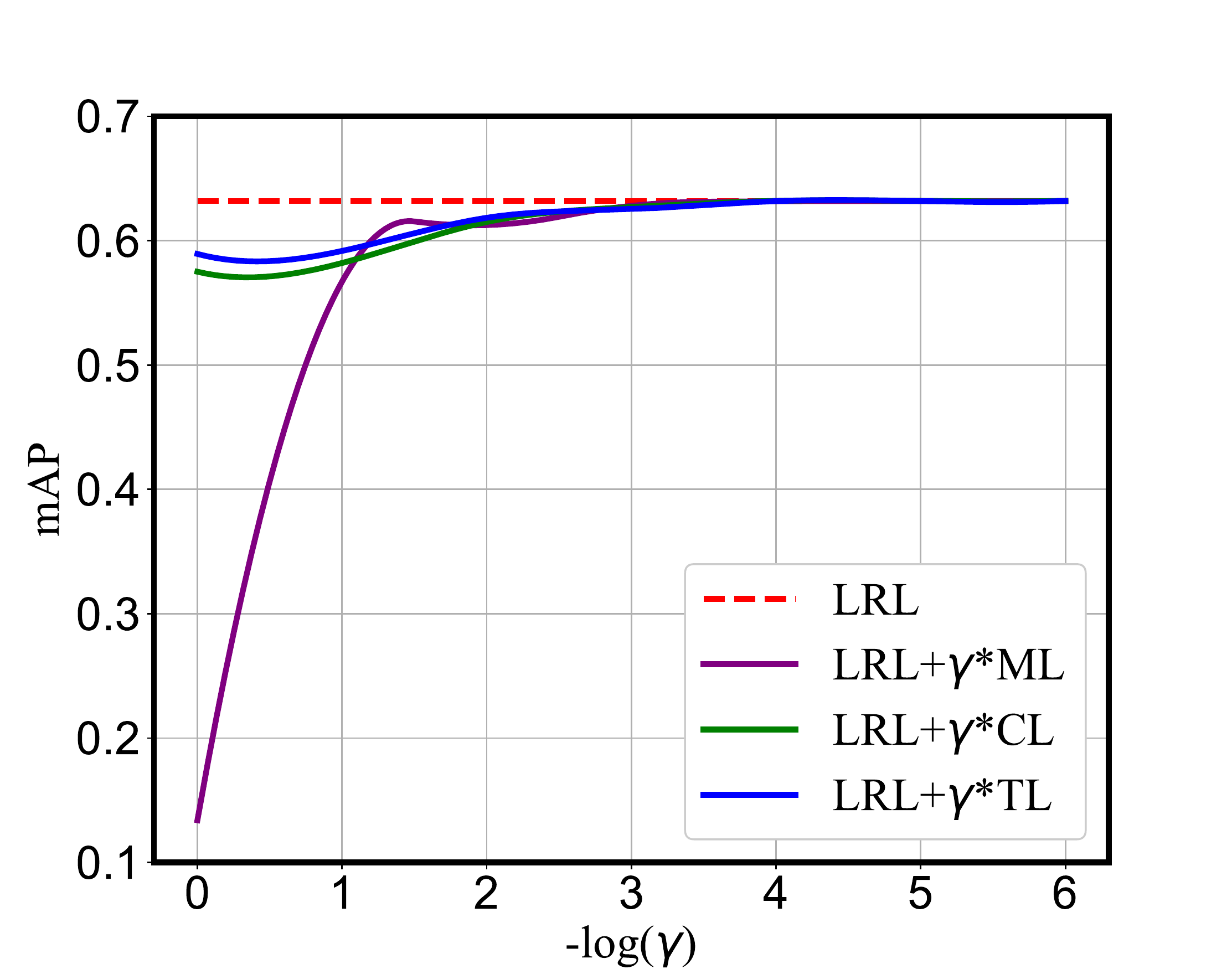}}}
		\subfigure[NUS-WIDE, CEL]{ 
			\raggedright{\includegraphics[width=0.261\textwidth]{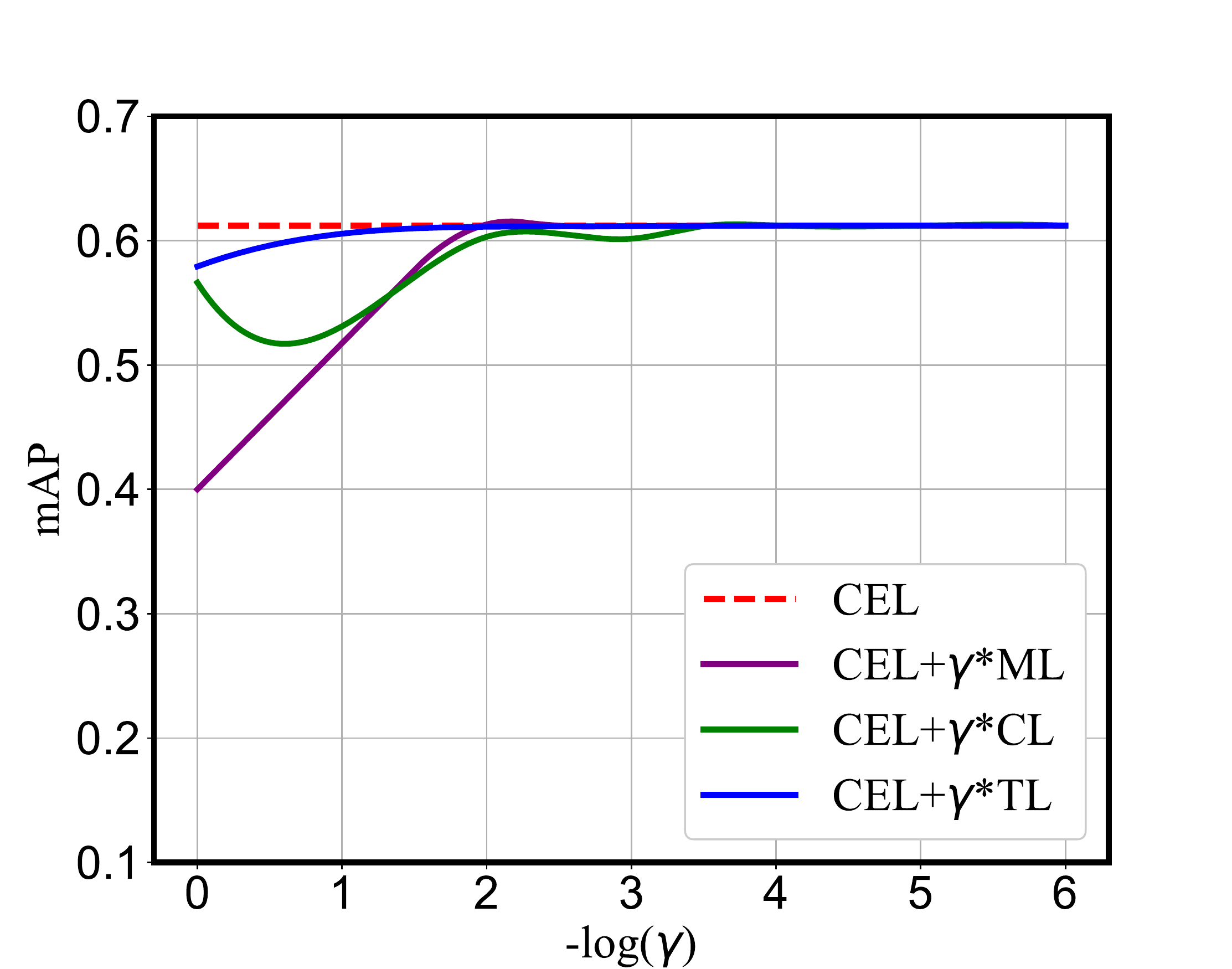}}}
		\subfigure[NUS-WIDE, PCL]{ 
			\raggedright{\includegraphics[width=0.261\textwidth]{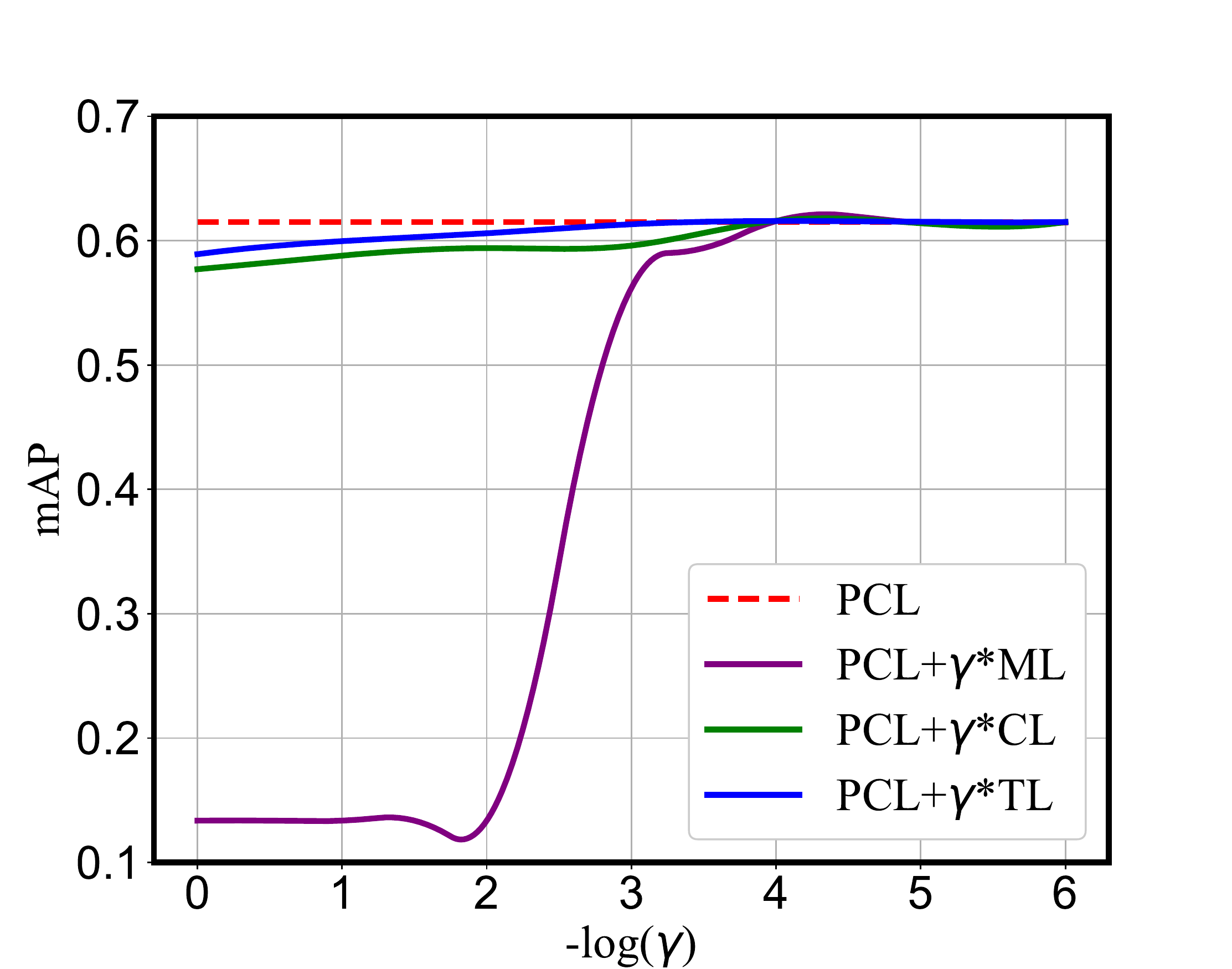}}}
		\caption{The impact of hybrid losses that combine class-wise and pair-wise losses. We show the average mAP values of text retrieval and image retrieval tasks under different combination weights $\gamma$. We compare the  performance of the hybrid loss combined by class-wise loss and different pair-wise losses in each sub-figure.}
		\label{multihead} 
	\end{flushleft}
\end{figure*}

\subsubsection{Evaluation Metrics}

The results of all the experiments are presented in terms of the mean average precision (mAP), which is the standard evaluation measure in cross-modal retrieval \cite{wang2015,wang2016comprehensive}. We compute the mAP scores for two different tasks: text retrieval using image query (I2T) and image retrieval using text query (T2I).
To calculate mAP, we first evaluate the average precision (AP) of a set of $R$ retrieved items by:
$AP = \frac{1}{T}\sum_{r=1}^{R}{P_r} \times \delta(r)$,
where T is the number of relevant items in the retrieved set, $P(r)$ represents the precision of the top $r$ retrieved items, and $\delta(r)$ is an indicator function, whose value is 1 if the $r$-th retrieved item is relevant (\emph{i.e.}, from the same class). The mAP scores are then calculated by averaging the AP values over all queries.

\subsubsection{Implementation Details} \label{sec:4.2}
The model architecture of CLIP4CMR is mainly based on CLIP, which consists of a visual encoder and a textual encoder that process image and text modalities separately. The visual encoder utilizes ResNet-50 \cite{he2016deep} as the base architecture, and makes several modifications to incorporate the style of Transformer \cite{vaswani2017attention}. Specifically, it adopts the modified version in ResNet-D \cite{he2019bag} and antialiased rect-2 blur pooling \cite{zhang2019making}, and then replaces the global average pooling layer with an attention pooling
mechanism. The attention pooling is implemented as a single layer of multi-head QKV attention where the query is conditioned on the pooled representation, and finally a $1024$-dimensional image representation is obtained.
The textual encoder first converts each token (including punctuation) of the input text into a lower-cased byte pair encoding (BPE) representation \cite{sennrich2015neural}, which is essentially a unique numeric ID. The vocabulary size in is $49,152$ and the text length is fixed as $77$ with the $[SOS]$ and $[EOS]$ tokens. Then the text IDs are mapped to $512$-dimensional word embeddings to be passed in the $12$-layer Transformer. Finally, the feature at the $[EOS]$ position is layer normalized and processed by a linear projection layer to generate $1024$-dimensional text representations.
Then we employ two fully connected layers to project the original image and text representations into a common representation space, respectively.
The entire network is optimized by Adam update rule \cite{kingma2014adam}. We set the initial learning rate to $10^{-4}$, the dropout ratio to $0.1$, the early stop to $20$, the batch size to $300$ and the maximal training epoch to $200$.

\noindent
\textbf{Hyper-parameter setting:} 
We report the results corresponding to the optimal hyper-parameters, where the dimension of the common representation space is $1,024$, and the scaling factor $\lambda$ in Eq.(\hyperref[eq8]{8}) is $1$. In addition, the margin $\Delta$ of pair-wise losses is set to be 0.2 as in most of previous work \cite{chen2020imram}. Further analysis of these hyper-parameters will be discussed in Section 4.4.

\begin{figure*}[h]\label{fig5}
	\begin{flushleft}		
		\centering 
		\subfigure[Wikipedia pair-wise]{ 
			\raggedright{\includegraphics[width=0.231\textwidth]{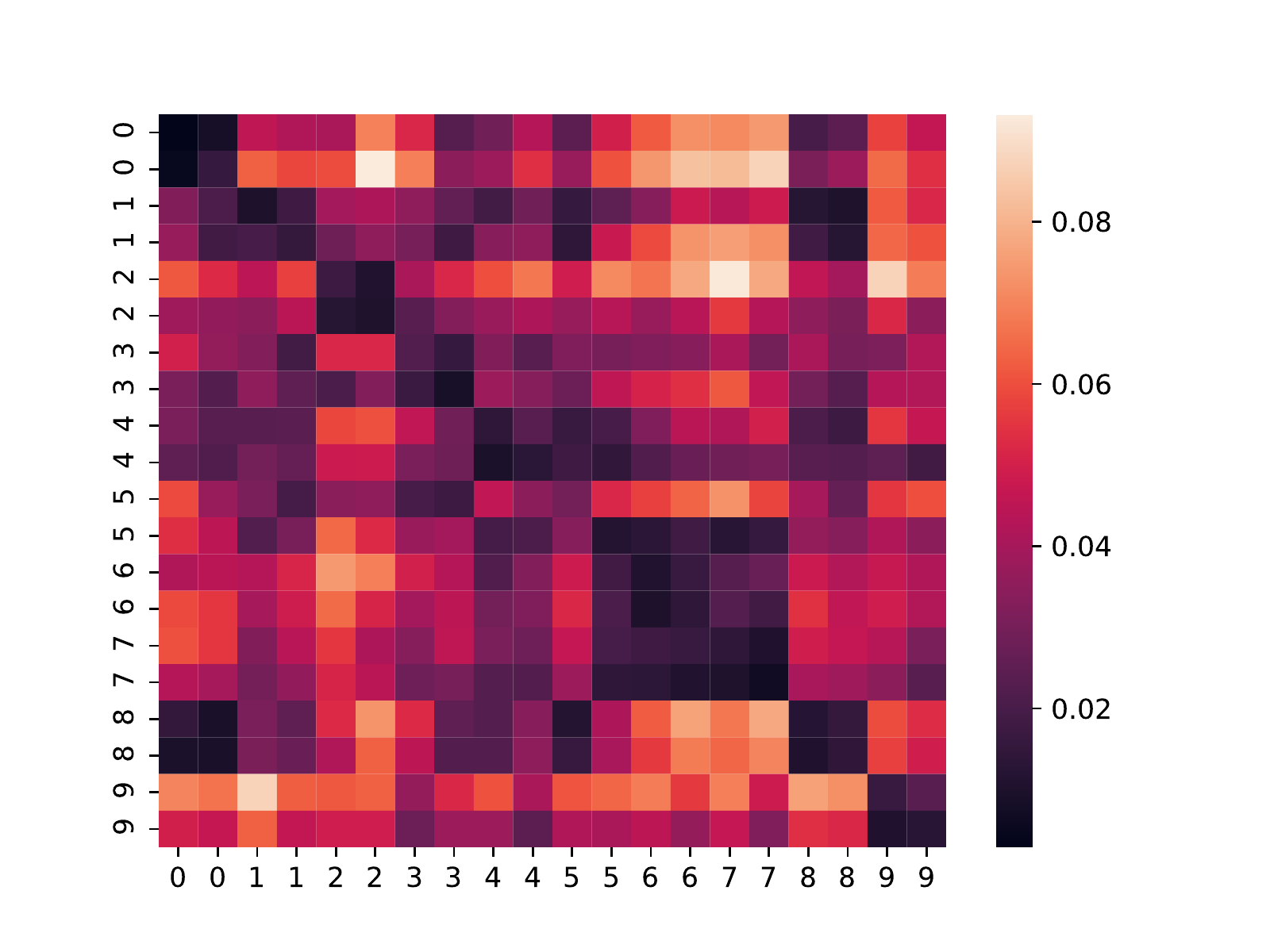}}}
		\subfigure[Pascal-Sentence pair-wise]{ 
			\raggedright{\includegraphics[width=0.231\textwidth]{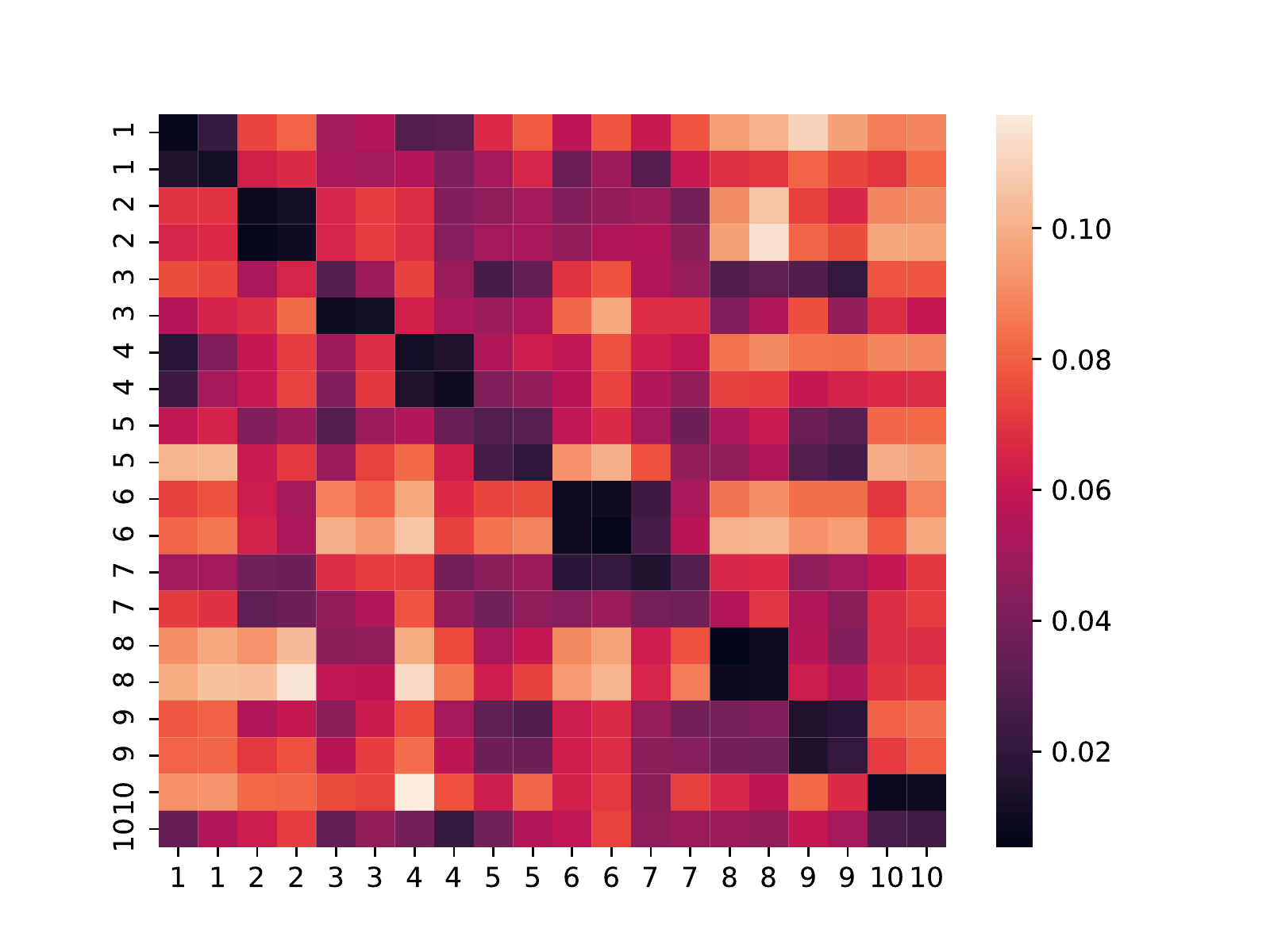}}}
		\subfigure[NUS-WIDE pair-wise]{ 
			\raggedright{\includegraphics[width=0.231\textwidth]{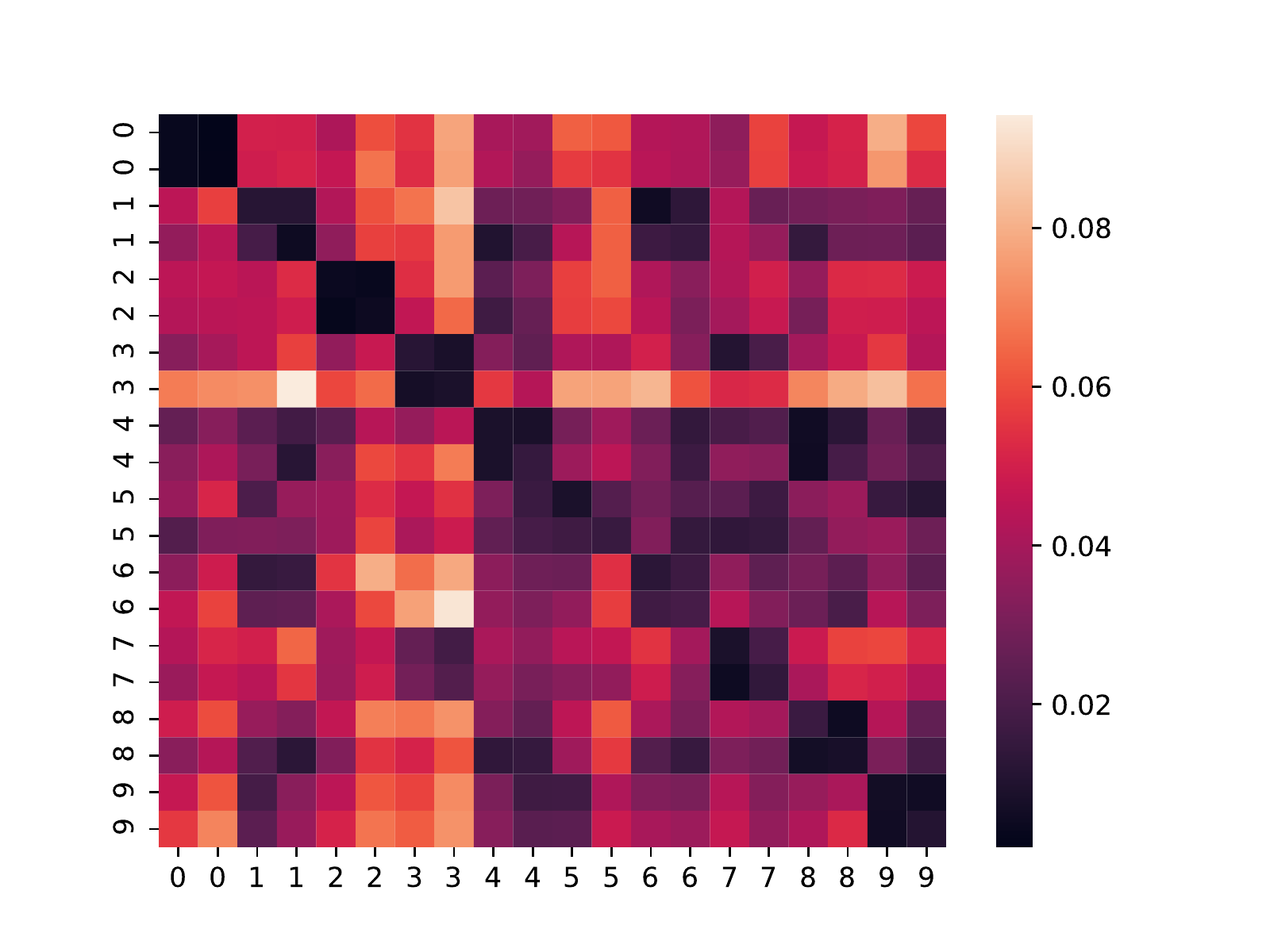}}}
		\subfigure[XmediaNet pair-wise]{ 
			\raggedright{\includegraphics[width=0.231\textwidth]{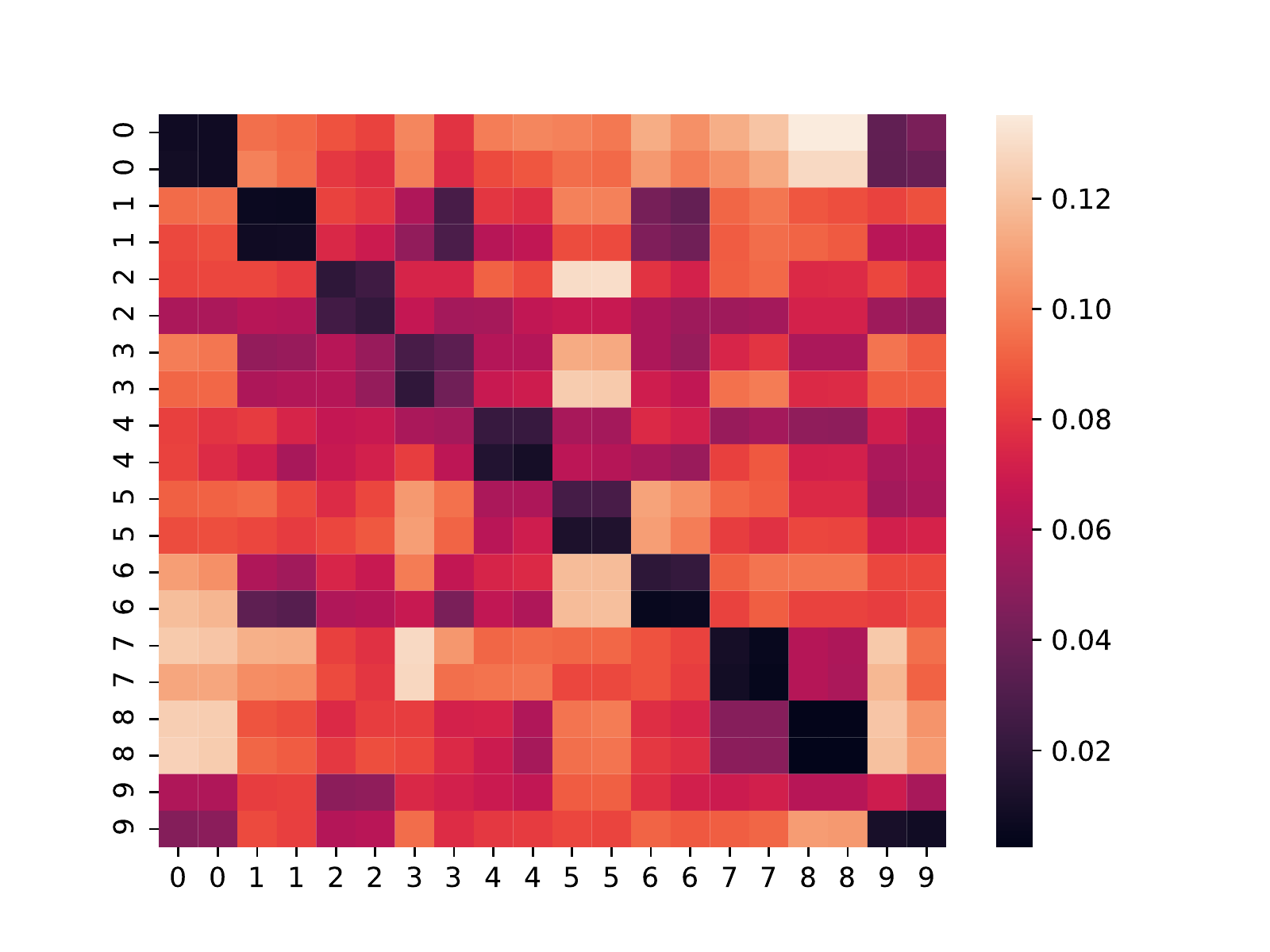}}}
		\subfigure[Wikipedia class-wise]{ 
			\includegraphics[width=0.231\textwidth]{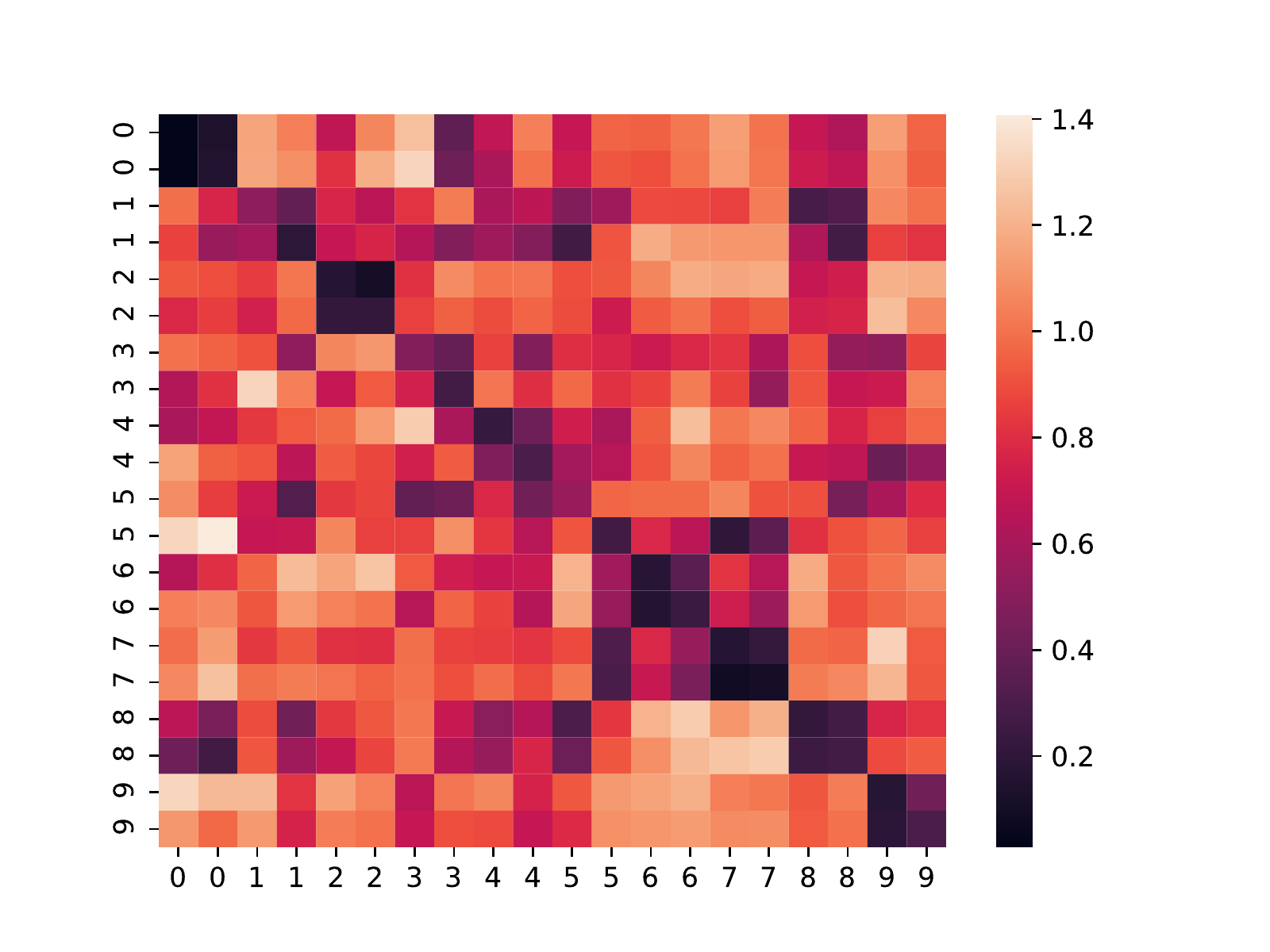}}		
		\subfigure[Pascal-Sentence class-wise]{ 
			\includegraphics[width=0.231\textwidth]{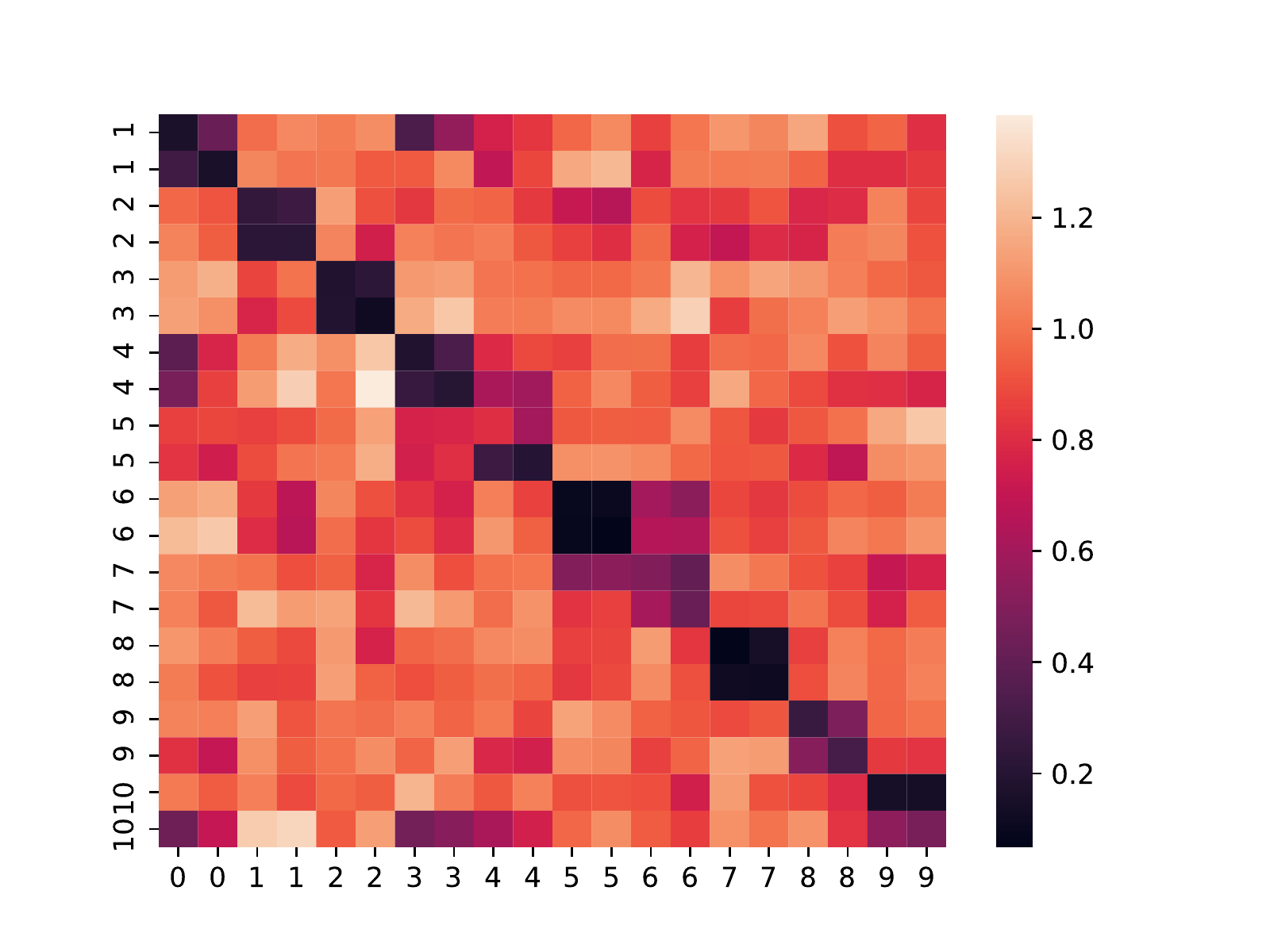}}		
		\subfigure[NUS-WIDE class-wise]{ 
			\includegraphics[width=0.231\textwidth]{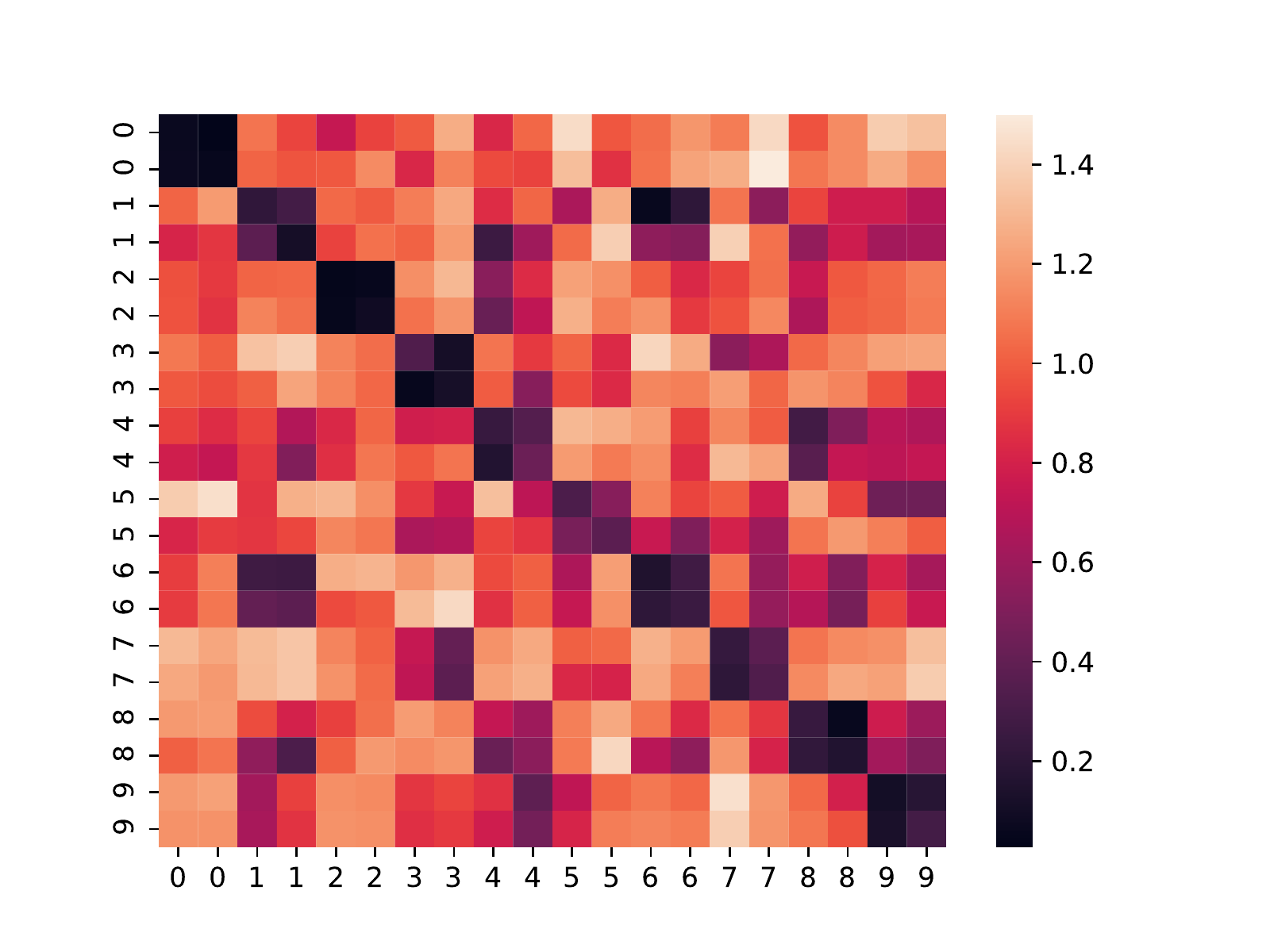}}
		\subfigure[XmediaNet class-wise]{ 
			\includegraphics[width=0.231\textwidth]{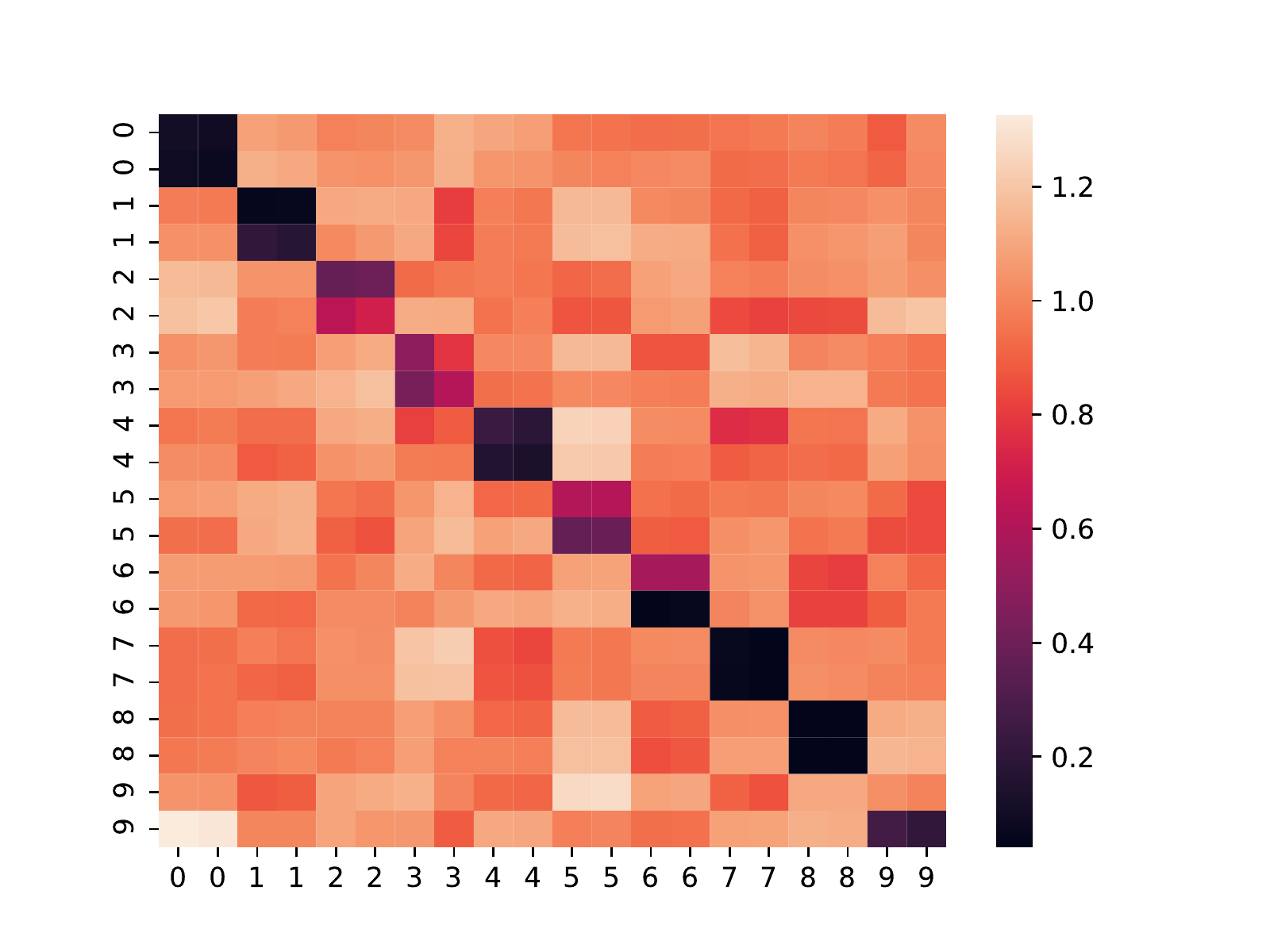}}
		\caption{Visualization of the distance matrix between the embeddings of test set learned by pair-wise loss and class-wise loss, respectively. X-axis denotes the image labels, and Y-axis denotes the text labels. }
		\label{multihead} 
	\end{flushleft}
\end{figure*}
\subsection{Study on the CLIP4CMR Performance}\label{sec:4.3}
\subsubsection{Comparative Results}
To evaluate the performance and impact of the vision-language pre-trained model CLIP in cross-modal retrieval, we compare the proposed CLIP4CMR with fourteen representative baseline methods, namely CCA \cite{hardoon2004canonical}, KCCA \cite{wang2015large},  Corr-AE \cite{feng2014cross}, JRL \cite{zhai2013learning}, CMDN \cite{peng2016cross}, JFSSL \cite{wang2015}, ACMR \cite{wang2017adversarial}, JLSLR \cite{wu2017joint}, MCSM \cite{peng2018modality}, CCL \cite{peng2017ccl}, CM-GANs \cite{peng2019cm}, DSCMR \cite{zhen2019deep}, PAN \cite{zeng2021pan} and MCCN \cite{zeng2021mccn}. Note that DSCMR and MCCN are two-stage methods, which use training data to train pre-classified visual and textual encoders followed by cross-modal retrieval. By this way, the two-stage training approach can significantly improve the performance of cross-modal retrieval in their original report, including the baseline methods.
Here we report the results of CLIP4CMR trained by prototype contrastive loss because of its overall better performance on all the datasets. We shall compare the performance of CLIP4CMR under different loss functions in Section 4.3. 

Table 1 reports the mAP scores of CLIP4CMR and the comparative methods. 
From the results, we can see that CLIP4CMR outperforms baseline methods on all benchmark datasets. Comparing with representative one-stage methods, our method outperforms PAN with the average mAP improvements 9.4\%, 0.7\%, 3.4\% and 8.7\% on Wikipedia, Pascal-Sentence, NUS-WIDE and XmediaNet, respectively. Moreover, CLIP4CMR still achieves better performance compared to recent two-stage methods, especially on the Wikipedia dataset with significant performance gains.
The promising results of CLIP4CMR indicate the superiority of CLIP in learning the visual and textual representations for boosting cross-modal retrieval.

\subsubsection{Visualization Analysis}
To further study how the superiority of CLIP4CMR is generated, we further examine the distributions of the intra-class image-text distances and inter-class image-text distances in the test set. Specifically, we collect $23,880$ intra-class image-text distances and $189,564$ inter-class image-text distances in the Wikipedia dataset, and $83,538$ intra-class image-text distances and $15,916,462$ inter-class image-text distances in the XmediaNet dataset. We adopt the previous SOTA method PAN \cite{zeng2021pan} for comparison, and show the visualization results in Figure 3. From the figure, we can see that the intra-class image-text distances of CLIP4CMR is obviously more compact than those of PAN, and the inter-class image-text distances of the two methods are not significantly different.
The visualization results show that the superiority of CLIP4CMR mainly comes from the more compact distribution of multimodal samples within class, which actually benefits from the prior knowledge of cross-modal semantic alignment obtained by the vision-language pre-trained model. 

\subsubsection{Summary and Implication for Future Research}
Benefited from the improvement of intra-class compactness, CLIP4CMR provides a promising baseline and can significantly facilitate cross-modal retrieval task.
This indicates that more future research efforts are needed to actively explore the effective utilization of powerful vision-language pre-trained models for cross-modal retrieval.

\begin{table*}[tbp]\label{tab3}
	\renewcommand\arraystretch{1.2}   
	\centering
	\caption{Average mAP scores (mean $\pm$ standard deviation) with imbalanced training data under the experimental settings of PAN \cite{zeng2021pan}.}
	\setlength{\tabcolsep}{2.1mm}{
		\begin{tabular}{ccccccccc}
			\hline
			\multirow{2}[2]{*}{percentage} & \multicolumn{4}{c}{Wikipedia} & \multicolumn{4}{c}{Pascal-Sentence}\\
			\cmidrule(r){2-5} 	\cmidrule(r){6-9}           & 	Baseline & DAVAE \cite{jing2020incomplete} & PAN \cite{zeng2021pan} & CLIP4CMR&	Baseline & DAVAE \cite{jing2020incomplete} & PAN \cite{zeng2021pan} & CLIP4CMR\\
			\hline		
			100\%I, 50\%T & 0.452$\pm$0.013 & 0.462$\pm$0.011 & 0.475$\pm$0.007&0.578$\pm$0.001 & 0.522$\pm$0.016 & 0.629$\pm$0.017 & 0.659$\pm$0.015 & 0.688$\pm$0.005 \\
			50\%I, 100\%T& 0.433$\pm$0.016 & 0.465$\pm$0.021 & 0.471$\pm$0.006 &0.573$\pm$0.003 & 0.548$\pm$0.019 & 0.618$\pm$0.014  & 0.652$\pm$0.008& 0.684$\pm$0.005 \\
			100\%I, 30\%T & 0.425$\pm$0.021 & 0.453$\pm$0.016 & 0.470$\pm$0.009 &0.571$\pm$0.003& 0.466$\pm$0.025 & 0.583$\pm$0.022 & 0.655$\pm$0.017& 0.687$\pm$0.004  \\
			30\%I, 100\%T & 0.417$\pm$0.019 & 0.448$\pm$0.018 & 0.462$\pm$0.010 &0.578$\pm$0.003&0.495$\pm$0.024 & 0.606$\pm$0.021 & 0.642$\pm$0.012 & 0.681$\pm$0.005 \\
			\hline
			100\%I, 100\%T & 0.482$\pm$0.003 & 0.485$\pm$0.006  & 0.489$\pm$0.002 &0.576$\pm$0.002 & 0.664$\pm$0.007  & 0.673$\pm$0.010  & 0.688$\pm$0.005& 0.690$\pm$0.003 \\
			\hline
	\end{tabular}}%
	\label{tab:addlabel}%
\end{table*}%
\begin{table*}[tbp]\label{tab3}
	\renewcommand\arraystretch{1.2}   
	\centering
	\caption{Average mAP scores (mean $\pm$ standard deviation) with extremely imbalanced training data.}
	\setlength{\tabcolsep}{2.1mm}{
		\begin{tabular}{ccccccccc}
			\hline
			\multirow{2}[2]{*}{percentage} & \multicolumn{4}{c}{Wikipedia} & \multicolumn{4}{c}{Pascal-Sentence}\\
			\cmidrule(r){2-5} 	\cmidrule(r){6-9}           & 	100\%I, 10\%T & 10\%I, 100\%T & 100\%I, 0\%T & 0\%I, 100\%T & 	100\%I, 10\%T & 10\%I, 100\%T & 100\%I, 0\%T & 0\%I, 100\%T\\	
			
			\hline
			CLIP4CMR &0.564$\pm$0.002 & 0.577$\pm$0.004  & 0.139$\pm$0.003 &0.129$\pm$0.005 & 0.682$\pm$0.003  & 0.673$\pm$0.006   & 0.100$\pm$0.014& 0.088$\pm$0.007\\
			\hline
	\end{tabular}}%
	\label{tab:addlabel}%
\end{table*}

\subsection{Study on the Design of Learning Objectives}\label{sec:4.3}

\subsubsection{Comparative Results}
To provide a fair comparison of the loss function design in the existing models, we deploy CLIP4CMR as the uniform framework as well as experimental tool for revisiting the most common pair-wise losses, class-wise losses and hybrid ones. Specifically, we unify the model architecture of CLIP4CMR, training protocol, parameter choice and random seed for a relatively objective comparison.
We compare three popular pair-wise losses namely modality-invariant loss (\emph{i.e.}, ML), contrastive loss (\emph{i.e.}, CL) and triplet loss (\emph{i.e.}, TL), as well as three popular class-wise losses namely linear regression loss (\emph{i.e.}, LRL), cross-entropy loss (\emph{i.e.}, CEL) and prototype contrastive loss (\emph{i.e.}, PCL).

Table 2 reports the performance comparison of different loss function design. From the results, we can see that the overall performance of the prototype contrastive loss on the four datasets is significantly better than the other loss functions, although its performance on Wikipedia and NUS-WIDE datasets is slightly lower than that of linear regression loss.
For the pair-wise losses, we can see that the performance of modality-invariant loss is very poor, which shows the necessity of considering negative samples for cross-modal retrieval.
Moreover, the results show that there is an obvious performance gap between pair-wise loss and class-wise loss. Specifically, the prototype contrastive loss outperforms the triplet loss with the average mAP improvements 4.0\%, 7.3\%, 1.6\% and 7.6\% on Wikipedia, Pascal-Sentence, NUS-WIDE and XmediaNet, respectively. 

Figure 4 illustrates the performance of the hybrid losses that combine class-wise and pair-wise losses.
We carefully compare nine hybrid losses under different combinations including LRL+$\gamma$ML, LRL+$\gamma$CL, LRL+$\gamma$TL, CEL+$\gamma$ML, CEL+$\gamma$CL, CEL+$\gamma$TL, PCL+$\gamma$ML, PCL+$\gamma$CL and PCL+$\gamma$TL, where $\gamma$ represents the combination weight. Since the combination weight $\gamma$ is a carefully selected parameter and the existing work does not provide a clear value, we tune the parameter $\gamma$ and show the average mAP values.
The results show that under all possible combinations, the hybrid losses of carefully adjusted parameter $\gamma$ have no obvious performance gains compared to applying class-wise loss alone.
This empirical finding is consistent with the perspective in the recently proposed method PAN \cite{zeng2021pan}, that is, a simple combination of pair-wise loss and class-wise loss in cross-modal retrieval may not be necessary.

\subsubsection{Visualization Analysis}
To further explore the reason for this obvious performance gap, we carry out a visualization experiment to analyze the difference of the common representation spaces obtained by pairwise loss and class-wise loss.
Concretely, we randomly select 20 image-text pairs from 10 classes of the test set, and each class evenly contains 2 image-text pairs. 
We choose triple loss and prototype contrastive loss as the representatives for pair-wise loss and class-wise loss respectively.
We illustrate the results in Figure 5, where the positions on the diagonal represent the intra-class image-text distances in the common representation space, and the other positions represent the inter-class image-text distances. 
The visualization results show that the inter-class distances in the common representation space obtained by the triplet loss are significantly smaller than that obtained by the prototype contrastive loss.
This indicates that there are a large number of negative sample pairs in the pair-wise loss that cannot be optimized, leading to poorer retrieval performance.
Therefore, simply combining pair-wise loss and class-wise loss does not guarantee the expected performance gains, and the performance of hybrid loss is better when the combination weight $\gamma$ is smaller, as shown in Figure 4.

\subsubsection{Summary and Implication for Future Research}
Under the unified experimental setting based on CLIP4CMR, the hybrid losses that combine pair-wise and class-wise losses have no obvious performance gains compared to applying the class-wise loss alone. 
This indicates that on the one hand, more future research efforts are needed to design effective  high-performing data-to-proxy relations in class-wise loss. On the other hand, the complementary research efforts to further explore the design of more fine-grained data-to-data relations in pair-wise loss (possibly by learning from the merits of class-wise loss) may also be needed.

\subsection{Study on Two Practical Issues}\label{sec:4.3}
To facilitate practical applications, we experiment on two key concerned issues here in practice: the robustness to modality imbalance and sensitivity to hyper-parameters.

\subsubsection{The Robustness to Modality Imbalance}
First, we follow the dataset split scheme in PAN \cite{zeng2021pan} to construct imbalanced training data, which includes two imbalanced ratios: retain 50\% text or image samples (\emph{i.e.}, 100\%I+50\%T or 50\%I+100\%T) and retain 30\% text or image samples (\emph{i.e.}, 100\%I+30\%T or 30\%I+100\%T). 
Then we further construct a more extreme imbalanced setting, that is, only 10\% text or image samples are retained (\emph{i.e.}, 100\%I+10\%T or 10\%I+100\%T). Finally, to show the importance of the coexistence of image and text modalities, we also compare the results of only retaining image samples (\emph{i.e.}, 100\%I+0\%T) and only retaining text samples (\emph{i.e.}, 0\%I+100\%T). For comparison, we compare with DAVAE \cite{jing2020incomplete}, PAN \cite{zeng2021pan}, and the baseline method of not processing imbalanced data. All compared results are reported in PAN.

Following PAN, we repeat each experiment five times and report the average mAP scores (mean $\pm$ standard deviation) in Table 3 and Table 4.
From the experimental results, we can see that the baseline method encounters an obvious performance decline in the face of modality imbalance, and the degree of performance decline is positively correlated with the proportion of modality imbalance. We can also see that DAEVE and PAN achieve significant performance improvements by reconstructing modality balanced data, validating the necessity of using modality balanced data during the training phase.  
However, the emergence of CLIP4CMR changes these previously formed perspectives. CLIP4CMR achieves significantly better performance under all the imbalanced settings, and it maintains slight performance degradation in some extremely imbalanced settings (\emph{i.e.}, 100\%I+10\%T and 10\%I+100\%T in Table 4). 
The robustness of CLIP4CMR shows that the image and text representations obtained by CLIP pre-trained on large-scale modality balanced data can greatly alleviate the imbalanced problem effortlessly, which is an important change brought by the vision-language pre-trained model for cross-modal retrieval.
In particular, the performance of the model drops seriously when we discard text or image samples (\emph{i.e.}, 100\%I+0\%T and 0\%I+100\%T in Table 4), indicating that image and text modalities coexist are important to modality imbalanced situation.

\begin{table}[tbp]
	\renewcommand\arraystretch{1.1}   
	\centering
	\caption{Parameter analysis of the dimension $d$.}
	\setlength{\tabcolsep}{0.75mm}{
		\begin{tabular}{c|cccc}
			\hline
			Parameter & Wikipedia & Pascal-Sentence & NUS-WIDE & XmediaNet \\
			\hline
			d=64  & 0.569  & 0.675  & 0.606  &  0.730\\
			d=128 & 0.576  & 0.687  & 0.609  & 0.738  \\
			d=256 & 0.582  & 0.691  & 0.613  & 0.743 \\
			d=512 & 0.583  & \textbf{0.695}  & 0.614  &0.748  \\
			d=1024 & \textbf{0.585}  & 0.694  & \textbf{0.615}  & \textbf{0.752}\\
			d=2048 & 0.581  & 0.694  & 0.615  & 0.750 \\
			\hline
	\end{tabular}}%
	\label{tab:addlabel}%
\end{table}%

\subsubsection{The Sensitivity to Hyper-parameters}
To investigate the influence of hyper-parameters on the retrieval performance, we examine the mAP values of CLIP4CMR by varying the dimensionality $d$ of the common representation space. Note that the previous work did not perform a detailed parameter analysis of the dimensionality $d$, but we believe this is necessary due to the importance of its value in analyzing the computational storage and time efficiency of cross-modal retrieval.
We vary $d$ from $32$ to $2048$, and show the impact of different values of $d$. We report the average mAP values of text retrieval (I2T) and image retrieval (T2I) tasks in Table 5. We can see that when $d=1024$, the overall performance of CLIP4CMR on the four datasets is the best. 
We can also see that the performance of CLIP4CMR decreases slightly when $d$ decreases, which means that CLIP4CMR can maintain considerable performance even in a more compact representation space.
In particular, CLIP4CMR can still maintain a small performance degradation in a very compact representation space (such as $d=64$), indicating that the retrieval model built on CLIP is almost insensitive to the dimension changes of the common representation space.
In addition, we also analyze the impact of the scaling factor $\lambda$ in the prototype contrastive loss. We vary $\lambda$ from 0.01 to 10 and show the impact in Figure 6. From the results, we can see that CLIP4CMR achieves the best average mAP value when $\lambda=1$, and the performance drops significantly when $\lambda=10$, suggesting that it is harder to train
larger scaling factors due to the numerical stability.

\begin{figure}[tbp]\label{fig5}
	\begin{flushleft}		
		\centering 
		\subfigure[Wikipedia]{ 
			\raggedright{\includegraphics[width=0.231\textwidth]{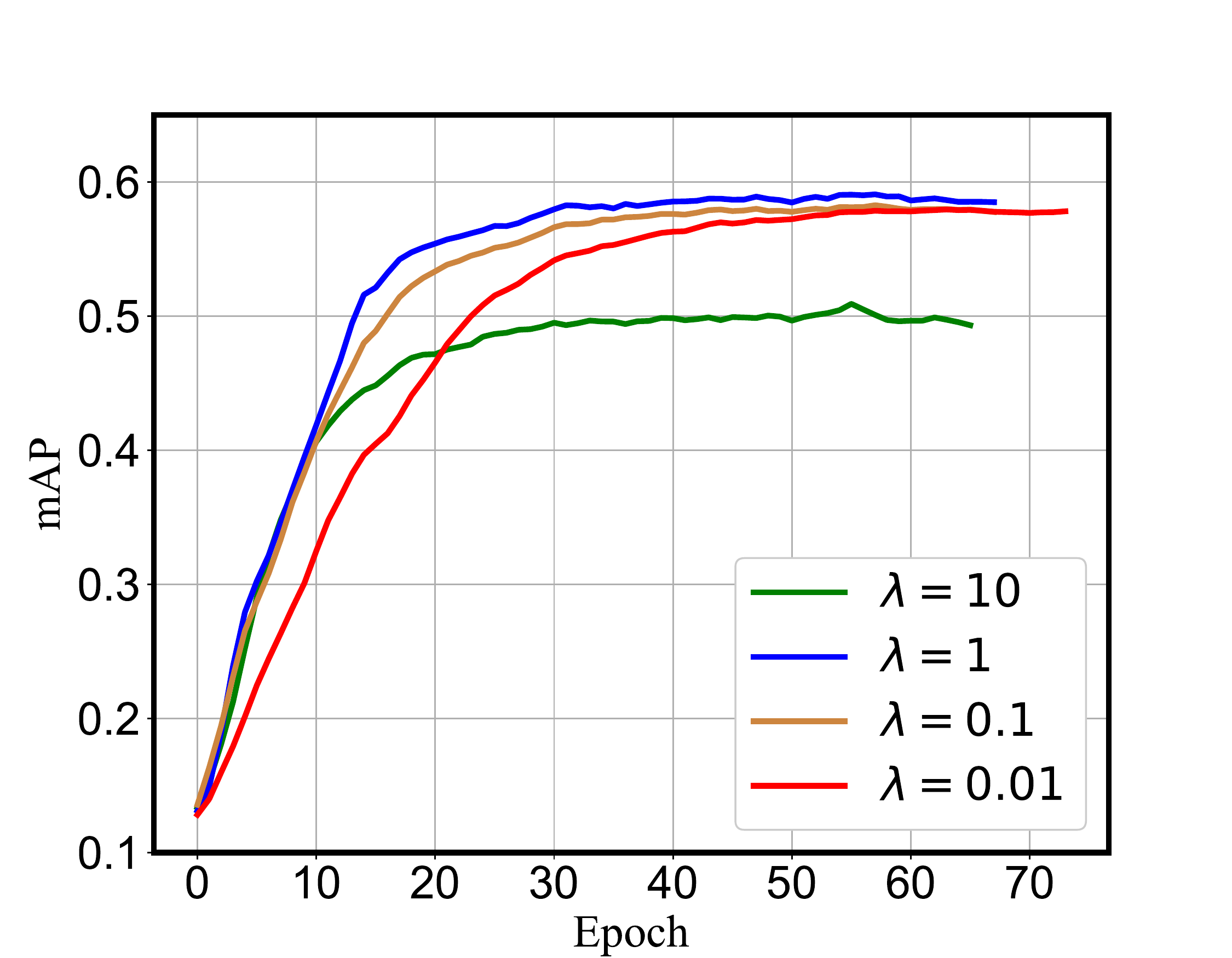}}}
		\subfigure[NUS-WIDE]{ 
			\raggedright{\includegraphics[width=0.231\textwidth]{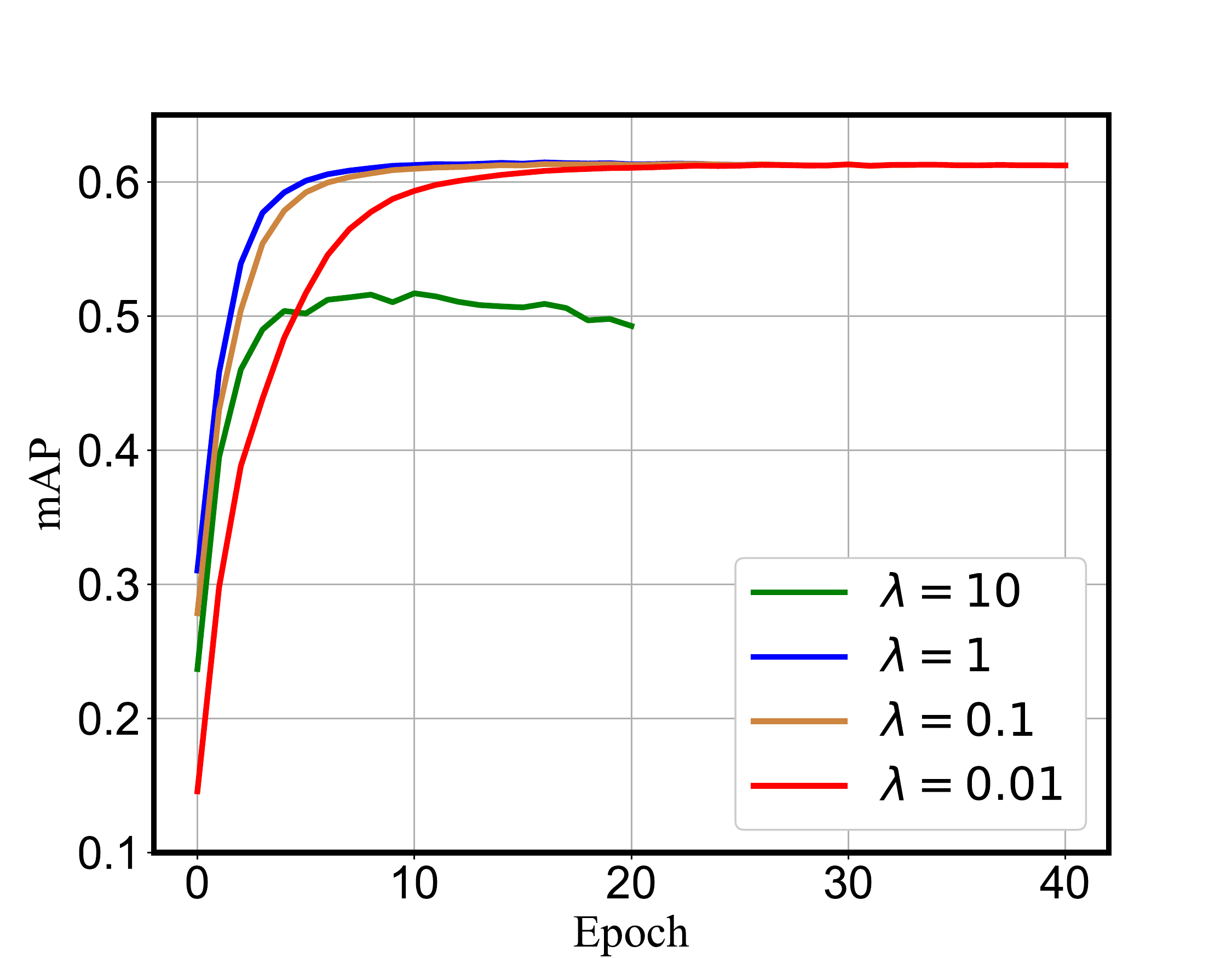}}}
		\caption{Parameter analysis of the scaling factor $\lambda$.} 
		\label{multihead} 
	\end{flushleft}
\end{figure}

\subsubsection{Summary and Implication for Future Research}
Cross-modal retrieval model built on CLIP markedly improves the robustness to modality imbalance and sensitivity to the dimension changes of the common representation space.
This indicates that with the help of vision-language pre-trained models, the dataset labeling and computational costs in practical applications can be greatly reduced in future research on cross-modal retrieval.

\section{Conclusion}
In this paper, we conduct a comprehensive empirical study to investigate the performance and impact of the pre-trained CLIP for cross-modal retrieval.
Our empirical study demonstrates that the CLIP4CMR framework built on CLIP can significantly facilitate the performance of cross-modal retrieval, together with the underlying rationale for this. The CLIP4CMR framework also provides a uniform experimental setting for the relatively objective comparison of the existing methods to gain valuable insights on loss function design.

\bibliographystyle{ACM-Reference-Format}
\bibliography{ref}


\begin{thebibliography}{64}


\ifx \showCODEN    \undefined \def \showCODEN     #1{\unskip}     \fi
\ifx \showDOI      \undefined \def \showDOI       #1{#1}\fi
\ifx \showISBNx    \undefined \def \showISBNx     #1{\unskip}     \fi
\ifx \showISBNxiii \undefined \def \showISBNxiii  #1{\unskip}     \fi
\ifx \showISSN     \undefined \def \showISSN      #1{\unskip}     \fi
\ifx \showLCCN     \undefined \def \showLCCN      #1{\unskip}     \fi
\ifx \shownote     \undefined \def \shownote      #1{#1}          \fi
\ifx \showarticletitle \undefined \def \showarticletitle #1{#1}   \fi
\ifx \showURL      \undefined \def \showURL       {\relax}        \fi
\providecommand\bibfield[2]{#2}
\providecommand\bibinfo[2]{#2}
\providecommand\natexlab[1]{#1}
\providecommand\showeprint[2][]{arXiv:#2}

\bibitem[\protect\citeauthoryear{Bellet, Habrard, and Sebban}{Bellet
  et~al\mbox{.}}{2013}]%
        {bellet2013survey}
\bibfield{author}{\bibinfo{person}{Aur{\'e}lien Bellet},
  \bibinfo{person}{Amaury Habrard}, {and} \bibinfo{person}{Marc Sebban}.}
  \bibinfo{year}{2013}\natexlab{}.
\newblock \showarticletitle{A survey on metric learning for feature vectors and
  structured data}.
\newblock \bibinfo{journal}{\emph{arXiv preprint arXiv:1306.6709}}
  (\bibinfo{year}{2013}).
\newblock


\bibitem[\protect\citeauthoryear{Cao, Gan, Cheng, Yu, Chen, and Liu}{Cao
  et~al\mbox{.}}{2020}]%
        {cao2020behind}
\bibfield{author}{\bibinfo{person}{Jize Cao}, \bibinfo{person}{Zhe Gan},
  \bibinfo{person}{Yu Cheng}, \bibinfo{person}{Licheng Yu},
  \bibinfo{person}{Yen-Chun Chen}, {and} \bibinfo{person}{Jingjing Liu}.}
  \bibinfo{year}{2020}\natexlab{}.
\newblock \showarticletitle{Behind the scene: Revealing the secrets of
  pre-trained vision-and-language models}. In
  \bibinfo{booktitle}{\emph{European Conference on Computer Vision}}. Springer,
  \bibinfo{pages}{565--580}.
\newblock


\bibitem[\protect\citeauthoryear{Carvalho, Cad{\`e}ne, Picard, Soulier, Thome,
  and Cord}{Carvalho et~al\mbox{.}}{2018}]%
        {carvalho2018cross}
\bibfield{author}{\bibinfo{person}{Micael Carvalho}, \bibinfo{person}{R{\'e}mi
  Cad{\`e}ne}, \bibinfo{person}{David Picard}, \bibinfo{person}{Laure Soulier},
  \bibinfo{person}{Nicolas Thome}, {and} \bibinfo{person}{Matthieu Cord}.}
  \bibinfo{year}{2018}\natexlab{}.
\newblock \showarticletitle{Cross-modal retrieval in the cooking context:
  Learning semantic text-image embeddings}. In \bibinfo{booktitle}{\emph{The
  41st International ACM SIGIR Conference on Research \& Development in
  Information Retrieval}}. \bibinfo{pages}{35--44}.
\newblock


\bibitem[\protect\citeauthoryear{Chen, Ding, Liu, Lin, Liu, and Han}{Chen
  et~al\mbox{.}}{2020}]%
        {chen2020imram}
\bibfield{author}{\bibinfo{person}{Hui Chen}, \bibinfo{person}{Guiguang Ding},
  \bibinfo{person}{Xudong Liu}, \bibinfo{person}{Zijia Lin},
  \bibinfo{person}{Ji Liu}, {and} \bibinfo{person}{Jungong Han}.}
  \bibinfo{year}{2020}\natexlab{}.
\newblock \showarticletitle{IMRAM: Iterative Matching with Recurrent Attention
  Memory for Cross-Modal Image-Text Retrieval}. In
  \bibinfo{booktitle}{\emph{Proceedings of the CVPR}}.
  \bibinfo{pages}{12655--12663}.
\newblock


\bibitem[\protect\citeauthoryear{Chen, Li, Yu, El~Kholy, Ahmed, Gan, Cheng, and
  Liu}{Chen et~al\mbox{.}}{2019}]%
        {chen2019uniter}
\bibfield{author}{\bibinfo{person}{Yen-Chun Chen}, \bibinfo{person}{Linjie Li},
  \bibinfo{person}{Licheng Yu}, \bibinfo{person}{Ahmed El~Kholy},
  \bibinfo{person}{Faisal Ahmed}, \bibinfo{person}{Zhe Gan},
  \bibinfo{person}{Yu Cheng}, {and} \bibinfo{person}{Jingjing Liu}.}
  \bibinfo{year}{2019}\natexlab{}.
\newblock \showarticletitle{Uniter: Learning universal image-text
  representations}.
\newblock  (\bibinfo{year}{2019}).
\newblock


\bibitem[\protect\citeauthoryear{Chua, Tang, Hong, Li, Luo, and Zheng}{Chua
  et~al\mbox{.}}{2009}]%
        {chua2009nus}
\bibfield{author}{\bibinfo{person}{Tat-Seng Chua}, \bibinfo{person}{Jinhui
  Tang}, \bibinfo{person}{Richang Hong}, \bibinfo{person}{Haojie Li},
  \bibinfo{person}{Zhiping Luo}, {and} \bibinfo{person}{Yantao Zheng}.}
  \bibinfo{year}{2009}\natexlab{}.
\newblock \showarticletitle{NUS-WIDE: a real-world web image database from
  National University of Singapore}. In \bibinfo{booktitle}{\emph{Proceedings
  of the CIVR}}. ACM, \bibinfo{pages}{1--9}.
\newblock


\bibitem[\protect\citeauthoryear{Chun, Oh, de~Rezende, Kalantidis, and
  Larlus}{Chun et~al\mbox{.}}{2021}]%
        {chun2021probabilistic}
\bibfield{author}{\bibinfo{person}{Sanghyuk Chun}, \bibinfo{person}{Seong~Joon
  Oh}, \bibinfo{person}{Rafael~Sampaio de Rezende}, \bibinfo{person}{Yannis
  Kalantidis}, {and} \bibinfo{person}{Diane Larlus}.}
  \bibinfo{year}{2021}\natexlab{}.
\newblock \showarticletitle{Probabilistic embeddings for cross-modal
  retrieval}. In \bibinfo{booktitle}{\emph{Proceedings of the CVPR}}.
  \bibinfo{pages}{8415--8424}.
\newblock


\bibitem[\protect\citeauthoryear{Desai and Johnson}{Desai and Johnson}{2021}]%
        {desai2021virtex}
\bibfield{author}{\bibinfo{person}{Karan Desai} {and} \bibinfo{person}{Justin
  Johnson}.} \bibinfo{year}{2021}\natexlab{}.
\newblock \showarticletitle{Virtex: Learning visual representations from
  textual annotations}. In \bibinfo{booktitle}{\emph{Proceedings of the
  IEEE/CVF Conference on Computer Vision and Pattern Recognition}}.
  \bibinfo{pages}{11162--11173}.
\newblock


\bibitem[\protect\citeauthoryear{Devlin, Chang, Lee, and Toutanova}{Devlin
  et~al\mbox{.}}{2018}]%
        {devlin2018bert}
\bibfield{author}{\bibinfo{person}{Jacob Devlin}, \bibinfo{person}{Ming-Wei
  Chang}, \bibinfo{person}{Kenton Lee}, {and} \bibinfo{person}{Kristina
  Toutanova}.} \bibinfo{year}{2018}\natexlab{}.
\newblock \showarticletitle{Bert: Pre-training of deep bidirectional
  transformers for language understanding}.
\newblock \bibinfo{journal}{\emph{arXiv preprint arXiv:1810.04805}}
  (\bibinfo{year}{2018}).
\newblock


\bibitem[\protect\citeauthoryear{Dosovitskiy, Beyer, Kolesnikov, Weissenborn,
  Zhai, Unterthiner, Dehghani, Minderer, Heigold, Gelly,
  et~al\mbox{.}}{Dosovitskiy et~al\mbox{.}}{2020}]%
        {dosovitskiy2020image}
\bibfield{author}{\bibinfo{person}{Alexey Dosovitskiy}, \bibinfo{person}{Lucas
  Beyer}, \bibinfo{person}{Alexander Kolesnikov}, \bibinfo{person}{Dirk
  Weissenborn}, \bibinfo{person}{Xiaohua Zhai}, \bibinfo{person}{Thomas
  Unterthiner}, \bibinfo{person}{Mostafa Dehghani}, \bibinfo{person}{Matthias
  Minderer}, \bibinfo{person}{Georg Heigold}, \bibinfo{person}{Sylvain Gelly},
  {et~al\mbox{.}}} \bibinfo{year}{2020}\natexlab{}.
\newblock \showarticletitle{An image is worth 16x16 words: Transformers for
  image recognition at scale}.
\newblock \bibinfo{journal}{\emph{arXiv preprint arXiv:2010.11929}}
  (\bibinfo{year}{2020}).
\newblock


\bibitem[\protect\citeauthoryear{Feng, Wang, and Li}{Feng
  et~al\mbox{.}}{2014}]%
        {feng2014cross}
\bibfield{author}{\bibinfo{person}{Fangxiang Feng}, \bibinfo{person}{Xiaojie
  Wang}, {and} \bibinfo{person}{Ruifan Li}.} \bibinfo{year}{2014}\natexlab{}.
\newblock \showarticletitle{Cross-modal retrieval with correspondence
  autoencoder}. In \bibinfo{booktitle}{\emph{Proceedings of the ACM MM}}. ACM,
  \bibinfo{pages}{7--16}.
\newblock


\bibitem[\protect\citeauthoryear{Geigle, Pfeiffer, Reimers, Vuli{\'c}, and
  Gurevych}{Geigle et~al\mbox{.}}{2021}]%
        {geigle2021retrieve}
\bibfield{author}{\bibinfo{person}{Gregor Geigle}, \bibinfo{person}{Jonas
  Pfeiffer}, \bibinfo{person}{Nils Reimers}, \bibinfo{person}{Ivan Vuli{\'c}},
  {and} \bibinfo{person}{Iryna Gurevych}.} \bibinfo{year}{2021}\natexlab{}.
\newblock \showarticletitle{Retrieve fast, rerank smart: Cooperative and joint
  approaches for improved cross-modal retrieval}.
\newblock \bibinfo{journal}{\emph{arXiv preprint arXiv:2103.11920}}
  (\bibinfo{year}{2021}).
\newblock


\bibitem[\protect\citeauthoryear{Hardoon, Szedmak, and Shawe-Taylor}{Hardoon
  et~al\mbox{.}}{2004}]%
        {hardoon2004canonical}
\bibfield{author}{\bibinfo{person}{David~R Hardoon}, \bibinfo{person}{Sandor
  Szedmak}, {and} \bibinfo{person}{John Shawe-Taylor}.}
  \bibinfo{year}{2004}\natexlab{}.
\newblock \showarticletitle{Canonical correlation analysis: An overview with
  application to learning methods}.
\newblock \bibinfo{journal}{\emph{Neural computation}} \bibinfo{volume}{16},
  \bibinfo{number}{12} (\bibinfo{year}{2004}), \bibinfo{pages}{2639--2664}.
\newblock


\bibitem[\protect\citeauthoryear{He, Zhang, Ren, and Sun}{He
  et~al\mbox{.}}{2016}]%
        {he2016deep}
\bibfield{author}{\bibinfo{person}{Kaiming He}, \bibinfo{person}{Xiangyu
  Zhang}, \bibinfo{person}{Shaoqing Ren}, {and} \bibinfo{person}{Jian Sun}.}
  \bibinfo{year}{2016}\natexlab{}.
\newblock \showarticletitle{Deep residual learning for image recognition}. In
  \bibinfo{booktitle}{\emph{Proceedings of the IEEE conference on computer
  vision and pattern recognition}}. \bibinfo{pages}{770--778}.
\newblock


\bibitem[\protect\citeauthoryear{He, Zhang, Zhang, Zhang, Xie, and Li}{He
  et~al\mbox{.}}{2019}]%
        {he2019bag}
\bibfield{author}{\bibinfo{person}{Tong He}, \bibinfo{person}{Zhi Zhang},
  \bibinfo{person}{Hang Zhang}, \bibinfo{person}{Zhongyue Zhang},
  \bibinfo{person}{Junyuan Xie}, {and} \bibinfo{person}{Mu Li}.}
  \bibinfo{year}{2019}\natexlab{}.
\newblock \showarticletitle{Bag of tricks for image classification with
  convolutional neural networks}. In \bibinfo{booktitle}{\emph{Proceedings of
  the IEEE/CVF Conference on Computer Vision and Pattern Recognition}}.
  \bibinfo{pages}{558--567}.
\newblock


\bibitem[\protect\citeauthoryear{Hendrycks and Gimpel}{Hendrycks and
  Gimpel}{2016}]%
        {hendrycks2016gaussian}
\bibfield{author}{\bibinfo{person}{Dan Hendrycks} {and} \bibinfo{person}{Kevin
  Gimpel}.} \bibinfo{year}{2016}\natexlab{}.
\newblock \showarticletitle{Gaussian error linear units (gelus)}.
\newblock \bibinfo{journal}{\emph{arXiv preprint arXiv:1606.08415}}
  (\bibinfo{year}{2016}).
\newblock


\bibitem[\protect\citeauthoryear{Hu, Lu, and Tan}{Hu et~al\mbox{.}}{2014}]%
        {hu2014discriminative}
\bibfield{author}{\bibinfo{person}{Junlin Hu}, \bibinfo{person}{Jiwen Lu},
  {and} \bibinfo{person}{Yap-Peng Tan}.} \bibinfo{year}{2014}\natexlab{}.
\newblock \showarticletitle{Discriminative deep metric learning for face
  verification in the wild}. In \bibinfo{booktitle}{\emph{Proceedings of the
  IEEE conference on computer vision and pattern recognition}}.
  \bibinfo{pages}{1875--1882}.
\newblock


\bibitem[\protect\citeauthoryear{Huang, Wang, and Wang}{Huang
  et~al\mbox{.}}{2017}]%
        {huang2017instance}
\bibfield{author}{\bibinfo{person}{Yan Huang}, \bibinfo{person}{Wei Wang},
  {and} \bibinfo{person}{Liang Wang}.} \bibinfo{year}{2017}\natexlab{}.
\newblock \showarticletitle{Instance-aware image and sentence matching with
  selective multimodal lstm}. In \bibinfo{booktitle}{\emph{Proceedings of the
  CVPR}}. \bibinfo{pages}{2310--2318}.
\newblock


\bibitem[\protect\citeauthoryear{Jing, Li, Zhu, Lu, Yang, and Huang}{Jing
  et~al\mbox{.}}{2020}]%
        {jing2020incomplete}
\bibfield{author}{\bibinfo{person}{Mengmeng Jing}, \bibinfo{person}{Jingjing
  Li}, \bibinfo{person}{Lei Zhu}, \bibinfo{person}{Ke Lu},
  \bibinfo{person}{Yang Yang}, {and} \bibinfo{person}{Zi Huang}.}
  \bibinfo{year}{2020}\natexlab{}.
\newblock \showarticletitle{Incomplete Cross-modal Retrieval with Dual-Aligned
  Variational Autoencoders}. In \bibinfo{booktitle}{\emph{Proceedings of the
  ACM international conference on Multimedia}}. \bibinfo{pages}{3283--3291}.
\newblock


\bibitem[\protect\citeauthoryear{Kim, Kim, Cho, and Kwak}{Kim
  et~al\mbox{.}}{2021}]%
        {kim2021embedding}
\bibfield{author}{\bibinfo{person}{Sungyeon Kim}, \bibinfo{person}{Dongwon
  Kim}, \bibinfo{person}{Minsu Cho}, {and} \bibinfo{person}{Suha Kwak}.}
  \bibinfo{year}{2021}\natexlab{}.
\newblock \showarticletitle{Embedding Transfer with Label Relaxation for
  Improved Metric Learning}. In \bibinfo{booktitle}{\emph{Proceedings of the
  IEEE/CVF Conference on Computer Vision and Pattern Recognition}}.
  \bibinfo{pages}{3967--3976}.
\newblock


\bibitem[\protect\citeauthoryear{Kingma and Ba}{Kingma and Ba}{2014}]%
        {kingma2014adam}
\bibfield{author}{\bibinfo{person}{Diederik~P Kingma} {and}
  \bibinfo{person}{Jimmy Ba}.} \bibinfo{year}{2014}\natexlab{}.
\newblock \showarticletitle{Adam: A method for stochastic optimization}.
\newblock \bibinfo{journal}{\emph{arXiv preprint arXiv:1412.6980}}
  (\bibinfo{year}{2014}).
\newblock


\bibitem[\protect\citeauthoryear{Li, Tang, Li, Xiao, Wu, Pu, and Zhuang}{Li
  et~al\mbox{.}}{2020a}]%
        {li2020topic}
\bibfield{author}{\bibinfo{person}{Jiacheng Li}, \bibinfo{person}{Siliang
  Tang}, \bibinfo{person}{Juncheng Li}, \bibinfo{person}{Jun Xiao},
  \bibinfo{person}{Fei Wu}, \bibinfo{person}{Shiliang Pu}, {and}
  \bibinfo{person}{Yueting Zhuang}.} \bibinfo{year}{2020}\natexlab{a}.
\newblock \showarticletitle{Topic Adaptation and Prototype Encoding for
  Few-Shot Visual Storytelling}.
\newblock \bibinfo{journal}{\emph{arXiv preprint arXiv:2008.04504}}
  (\bibinfo{year}{2020}).
\newblock


\bibitem[\protect\citeauthoryear{Li, Yin, Li, Zhang, Hu, Zhang, Wang, Hu, Dong,
  Wei, et~al\mbox{.}}{Li et~al\mbox{.}}{2020b}]%
        {li2020oscar}
\bibfield{author}{\bibinfo{person}{Xiujun Li}, \bibinfo{person}{Xi Yin},
  \bibinfo{person}{Chunyuan Li}, \bibinfo{person}{Pengchuan Zhang},
  \bibinfo{person}{Xiaowei Hu}, \bibinfo{person}{Lei Zhang},
  \bibinfo{person}{Lijuan Wang}, \bibinfo{person}{Houdong Hu},
  \bibinfo{person}{Li Dong}, \bibinfo{person}{Furu Wei}, {et~al\mbox{.}}}
  \bibinfo{year}{2020}\natexlab{b}.
\newblock \showarticletitle{Oscar: Object-semantics aligned pre-training for
  vision-language tasks}. In \bibinfo{booktitle}{\emph{European Conference on
  Computer Vision}}. Springer, \bibinfo{pages}{121--137}.
\newblock


\bibitem[\protect\citeauthoryear{Liu, Ott, Goyal, Du, Joshi, Chen, Levy, Lewis,
  Zettlemoyer, and Stoyanov}{Liu et~al\mbox{.}}{2019}]%
        {liu2019roberta}
\bibfield{author}{\bibinfo{person}{Yinhan Liu}, \bibinfo{person}{Myle Ott},
  \bibinfo{person}{Naman Goyal}, \bibinfo{person}{Jingfei Du},
  \bibinfo{person}{Mandar Joshi}, \bibinfo{person}{Danqi Chen},
  \bibinfo{person}{Omer Levy}, \bibinfo{person}{Mike Lewis},
  \bibinfo{person}{Luke Zettlemoyer}, {and} \bibinfo{person}{Veselin
  Stoyanov}.} \bibinfo{year}{2019}\natexlab{}.
\newblock \showarticletitle{Roberta: A robustly optimized bert pretraining
  approach}.
\newblock \bibinfo{journal}{\emph{arXiv preprint arXiv:1907.11692}}
  (\bibinfo{year}{2019}).
\newblock


\bibitem[\protect\citeauthoryear{Lu, Batra, Parikh, and Lee}{Lu
  et~al\mbox{.}}{2019}]%
        {lu2019vilbert}
\bibfield{author}{\bibinfo{person}{Jiasen Lu}, \bibinfo{person}{Dhruv Batra},
  \bibinfo{person}{Devi Parikh}, {and} \bibinfo{person}{Stefan Lee}.}
  \bibinfo{year}{2019}\natexlab{}.
\newblock \showarticletitle{Vilbert: Pretraining task-agnostic visiolinguistic
  representations for vision-and-language tasks}.
\newblock \bibinfo{journal}{\emph{arXiv preprint arXiv:1908.02265}}
  (\bibinfo{year}{2019}).
\newblock


\bibitem[\protect\citeauthoryear{Luo, Ji, Zhong, Chen, Lei, Duan, and Li}{Luo
  et~al\mbox{.}}{2021}]%
        {luo2021clip4clip}
\bibfield{author}{\bibinfo{person}{Huaishao Luo}, \bibinfo{person}{Lei Ji},
  \bibinfo{person}{Ming Zhong}, \bibinfo{person}{Yang Chen},
  \bibinfo{person}{Wen Lei}, \bibinfo{person}{Nan Duan}, {and}
  \bibinfo{person}{Tianrui Li}.} \bibinfo{year}{2021}\natexlab{}.
\newblock \showarticletitle{Clip4clip: An empirical study of clip for end to
  end video clip retrieval}.
\newblock \bibinfo{journal}{\emph{arXiv preprint arXiv:2104.08860}}
  (\bibinfo{year}{2021}).
\newblock


\bibitem[\protect\citeauthoryear{Peng, Huang, and Qi}{Peng
  et~al\mbox{.}}{2016}]%
        {peng2016cross}
\bibfield{author}{\bibinfo{person}{Yuxin Peng}, \bibinfo{person}{Xin Huang},
  {and} \bibinfo{person}{Jinwei Qi}.} \bibinfo{year}{2016}\natexlab{}.
\newblock \showarticletitle{Cross-media shared representation by hierarchical
  learning with multiple deep networks.}. In
  \bibinfo{booktitle}{\emph{Proceedings of the IJCAI}}.
  \bibinfo{pages}{3846--3853}.
\newblock


\bibitem[\protect\citeauthoryear{Peng and Qi}{Peng and Qi}{2019}]%
        {peng2019cm}
\bibfield{author}{\bibinfo{person}{Yuxin Peng} {and} \bibinfo{person}{Jinwei
  Qi}.} \bibinfo{year}{2019}\natexlab{}.
\newblock \showarticletitle{CM-GANs: Cross-modal generative adversarial
  networks for common representation learning}.
\newblock \bibinfo{journal}{\emph{Transactions on Multimedia Computing,
  Communications, and Applications}} \bibinfo{volume}{15}, \bibinfo{number}{1}
  (\bibinfo{year}{2019}), \bibinfo{pages}{1--24}.
\newblock


\bibitem[\protect\citeauthoryear{Peng, Qi, Huang, and Yuan}{Peng
  et~al\mbox{.}}{2017}]%
        {peng2017ccl}
\bibfield{author}{\bibinfo{person}{Yuxin Peng}, \bibinfo{person}{Jinwei Qi},
  \bibinfo{person}{Xin Huang}, {and} \bibinfo{person}{Yuxin Yuan}.}
  \bibinfo{year}{2017}\natexlab{}.
\newblock \showarticletitle{CCL: Cross-modal correlation learning with
  multigrained fusion by hierarchical network}.
\newblock \bibinfo{journal}{\emph{IEEE Transactions on Multimedia}}
  \bibinfo{volume}{20}, \bibinfo{number}{2} (\bibinfo{year}{2017}),
  \bibinfo{pages}{405--420}.
\newblock


\bibitem[\protect\citeauthoryear{Peng, Qi, and Yuan}{Peng
  et~al\mbox{.}}{2018}]%
        {peng2018modality}
\bibfield{author}{\bibinfo{person}{Yuxin Peng}, \bibinfo{person}{Jinwei Qi},
  {and} \bibinfo{person}{Yuxin Yuan}.} \bibinfo{year}{2018}\natexlab{}.
\newblock \showarticletitle{Modality-specific cross-modal similarity
  measurement with recurrent attention network}.
\newblock \bibinfo{journal}{\emph{IEEE Transactions on Image Processing}}
  \bibinfo{volume}{27}, \bibinfo{number}{11} (\bibinfo{year}{2018}),
  \bibinfo{pages}{5585--5599}.
\newblock


\bibitem[\protect\citeauthoryear{Radford, Kim, Hallacy, Ramesh, Goh, Agarwal,
  Sastry, Askell, Mishkin, Clark, et~al\mbox{.}}{Radford et~al\mbox{.}}{2021}]%
        {radford2021learning}
\bibfield{author}{\bibinfo{person}{Alec Radford}, \bibinfo{person}{Jong~Wook
  Kim}, \bibinfo{person}{Chris Hallacy}, \bibinfo{person}{Aditya Ramesh},
  \bibinfo{person}{Gabriel Goh}, \bibinfo{person}{Sandhini Agarwal},
  \bibinfo{person}{Girish Sastry}, \bibinfo{person}{Amanda Askell},
  \bibinfo{person}{Pamela Mishkin}, \bibinfo{person}{Jack Clark},
  {et~al\mbox{.}}} \bibinfo{year}{2021}\natexlab{}.
\newblock \showarticletitle{Learning transferable visual models from natural
  language supervision}.
\newblock \bibinfo{journal}{\emph{arXiv preprint arXiv:2103.00020}}
  (\bibinfo{year}{2021}).
\newblock


\bibitem[\protect\citeauthoryear{Radford, Wu, Child, Luan, Amodei, Sutskever,
  et~al\mbox{.}}{Radford et~al\mbox{.}}{2019}]%
        {radford2019language}
\bibfield{author}{\bibinfo{person}{Alec Radford}, \bibinfo{person}{Jeffrey Wu},
  \bibinfo{person}{Rewon Child}, \bibinfo{person}{David Luan},
  \bibinfo{person}{Dario Amodei}, \bibinfo{person}{Ilya Sutskever},
  {et~al\mbox{.}}} \bibinfo{year}{2019}\natexlab{}.
\newblock \showarticletitle{Language models are unsupervised multitask
  learners}.
\newblock \bibinfo{journal}{\emph{OpenAI blog}} \bibinfo{volume}{1},
  \bibinfo{number}{8} (\bibinfo{year}{2019}), \bibinfo{pages}{9}.
\newblock


\bibitem[\protect\citeauthoryear{Rashtchian, Young, Hodosh, and
  Hockenmaier}{Rashtchian et~al\mbox{.}}{2010}]%
        {rashtchian2010collecting}
\bibfield{author}{\bibinfo{person}{Cyrus Rashtchian}, \bibinfo{person}{Peter
  Young}, \bibinfo{person}{Micah Hodosh}, {and} \bibinfo{person}{Julia
  Hockenmaier}.} \bibinfo{year}{2010}\natexlab{}.
\newblock \showarticletitle{Collecting image annotations using amazon’s
  mechanical turk}. In \bibinfo{booktitle}{\emph{Proceedings of the NAACL}}.
  \bibinfo{pages}{139--147}.
\newblock


\bibitem[\protect\citeauthoryear{Rasiwasia, Costa~Pereira, Coviello, Doyle,
  Lanckriet, Levy, and Vasconcelos}{Rasiwasia et~al\mbox{.}}{2010}]%
        {rasiwasia2010new}
\bibfield{author}{\bibinfo{person}{Nikhil Rasiwasia}, \bibinfo{person}{Jose
  Costa~Pereira}, \bibinfo{person}{Emanuele Coviello}, \bibinfo{person}{Gabriel
  Doyle}, \bibinfo{person}{Gert~RG Lanckriet}, \bibinfo{person}{Roger Levy},
  {and} \bibinfo{person}{Nuno Vasconcelos}.} \bibinfo{year}{2010}\natexlab{}.
\newblock \showarticletitle{A new approach to cross-modal multimedia
  retrieval}. In \bibinfo{booktitle}{\emph{Proceedings of the ACM MM}}.
  \bibinfo{pages}{251--260}.
\newblock


\bibitem[\protect\citeauthoryear{Roth, Milbich, Sinha, Gupta, Ommer, and
  Cohen}{Roth et~al\mbox{.}}{2020}]%
        {roth2020revisiting}
\bibfield{author}{\bibinfo{person}{Karsten Roth}, \bibinfo{person}{Timo
  Milbich}, \bibinfo{person}{Samarth Sinha}, \bibinfo{person}{Prateek Gupta},
  \bibinfo{person}{Bjorn Ommer}, {and} \bibinfo{person}{Joseph~Paul Cohen}.}
  \bibinfo{year}{2020}\natexlab{}.
\newblock \showarticletitle{Revisiting training strategies and generalization
  performance in deep metric learning}. In
  \bibinfo{booktitle}{\emph{International Conference on Machine Learning}}.
  PMLR, \bibinfo{pages}{8242--8252}.
\newblock


\bibitem[\protect\citeauthoryear{Sennrich, Haddow, and Birch}{Sennrich
  et~al\mbox{.}}{2015}]%
        {sennrich2015neural}
\bibfield{author}{\bibinfo{person}{Rico Sennrich}, \bibinfo{person}{Barry
  Haddow}, {and} \bibinfo{person}{Alexandra Birch}.}
  \bibinfo{year}{2015}\natexlab{}.
\newblock \showarticletitle{Neural machine translation of rare words with
  subword units}.
\newblock \bibinfo{journal}{\emph{arXiv preprint arXiv:1508.07909}}
  (\bibinfo{year}{2015}).
\newblock


\bibitem[\protect\citeauthoryear{Shen, Li, Tan, Bansal, Rohrbach, Chang, Yao,
  and Keutzer}{Shen et~al\mbox{.}}{2021}]%
        {shen2021much}
\bibfield{author}{\bibinfo{person}{Sheng Shen}, \bibinfo{person}{Liunian~Harold
  Li}, \bibinfo{person}{Hao Tan}, \bibinfo{person}{Mohit Bansal},
  \bibinfo{person}{Anna Rohrbach}, \bibinfo{person}{Kai-Wei Chang},
  \bibinfo{person}{Zhewei Yao}, {and} \bibinfo{person}{Kurt Keutzer}.}
  \bibinfo{year}{2021}\natexlab{}.
\newblock \showarticletitle{How Much Can CLIP Benefit Vision-and-Language
  Tasks?}
\newblock \bibinfo{journal}{\emph{arXiv preprint arXiv:2107.06383}}
  (\bibinfo{year}{2021}).
\newblock


\bibitem[\protect\citeauthoryear{Shin, Ishii, and Narihira}{Shin
  et~al\mbox{.}}{2021}]%
        {shin2021perspectives}
\bibfield{author}{\bibinfo{person}{Andrew Shin}, \bibinfo{person}{Masato
  Ishii}, {and} \bibinfo{person}{Takuya Narihira}.}
  \bibinfo{year}{2021}\natexlab{}.
\newblock \showarticletitle{Perspectives and Prospects on Transformer
  Architecture for Cross-Modal Tasks with Language and Vision}.
\newblock \bibinfo{journal}{\emph{arXiv preprint arXiv:2103.04037}}
  (\bibinfo{year}{2021}).
\newblock


\bibitem[\protect\citeauthoryear{Song and Soleymani}{Song and
  Soleymani}{2019}]%
        {song2019polysemous}
\bibfield{author}{\bibinfo{person}{Yale Song} {and} \bibinfo{person}{Mohammad
  Soleymani}.} \bibinfo{year}{2019}\natexlab{}.
\newblock \showarticletitle{Polysemous visual-semantic embedding for
  cross-modal retrieval}. In \bibinfo{booktitle}{\emph{Proceedings of the
  CVPR}}. \bibinfo{pages}{1979--1988}.
\newblock


\bibitem[\protect\citeauthoryear{Su, Zhu, Cao, Li, Lu, Wei, and Dai}{Su
  et~al\mbox{.}}{2019}]%
        {su2019vl}
\bibfield{author}{\bibinfo{person}{Weijie Su}, \bibinfo{person}{Xizhou Zhu},
  \bibinfo{person}{Yue Cao}, \bibinfo{person}{Bin Li}, \bibinfo{person}{Lewei
  Lu}, \bibinfo{person}{Furu Wei}, {and} \bibinfo{person}{Jifeng Dai}.}
  \bibinfo{year}{2019}\natexlab{}.
\newblock \showarticletitle{Vl-bert: Pre-training of generic visual-linguistic
  representations}.
\newblock \bibinfo{journal}{\emph{arXiv preprint arXiv:1908.08530}}
  (\bibinfo{year}{2019}).
\newblock


\bibitem[\protect\citeauthoryear{Sun, Chen, Li, Wang, Fang, and Liu}{Sun
  et~al\mbox{.}}{2021}]%
        {sun2021lightningdot}
\bibfield{author}{\bibinfo{person}{Siqi Sun}, \bibinfo{person}{Yen-Chun Chen},
  \bibinfo{person}{Linjie Li}, \bibinfo{person}{Shuohang Wang},
  \bibinfo{person}{Yuwei Fang}, {and} \bibinfo{person}{Jingjing Liu}.}
  \bibinfo{year}{2021}\natexlab{}.
\newblock \showarticletitle{LightningDOT: Pre-training Visual-Semantic
  Embeddings for Real-Time Image-Text Retrieval}. In
  \bibinfo{booktitle}{\emph{Proceedings of the 2021 Conference of the North
  American Chapter of the Association for Computational Linguistics: Human
  Language Technologies}}. \bibinfo{pages}{982--997}.
\newblock


\bibitem[\protect\citeauthoryear{Tan and Bansal}{Tan and Bansal}{2019}]%
        {tan2019lxmert}
\bibfield{author}{\bibinfo{person}{Hao Tan} {and} \bibinfo{person}{Mohit
  Bansal}.} \bibinfo{year}{2019}\natexlab{}.
\newblock \showarticletitle{Lxmert: Learning cross-modality encoder
  representations from transformers}.
\newblock \bibinfo{journal}{\emph{arXiv preprint arXiv:1908.07490}}
  (\bibinfo{year}{2019}).
\newblock


\bibitem[\protect\citeauthoryear{Vaswani, Shazeer, Parmar, Uszkoreit, Jones,
  Gomez, Kaiser, and Polosukhin}{Vaswani et~al\mbox{.}}{2017}]%
        {vaswani2017attention}
\bibfield{author}{\bibinfo{person}{Ashish Vaswani}, \bibinfo{person}{Noam
  Shazeer}, \bibinfo{person}{Niki Parmar}, \bibinfo{person}{Jakob Uszkoreit},
  \bibinfo{person}{Llion Jones}, \bibinfo{person}{Aidan~N Gomez},
  \bibinfo{person}{{\L}ukasz Kaiser}, {and} \bibinfo{person}{Illia
  Polosukhin}.} \bibinfo{year}{2017}\natexlab{}.
\newblock \showarticletitle{Attention is all you need}. In
  \bibinfo{booktitle}{\emph{Advances in neural information processing
  systems}}. \bibinfo{pages}{5998--6008}.
\newblock


\bibitem[\protect\citeauthoryear{Vinyals, Toshev, Bengio, and Erhan}{Vinyals
  et~al\mbox{.}}{2015}]%
        {vinyals2015show}
\bibfield{author}{\bibinfo{person}{Oriol Vinyals}, \bibinfo{person}{Alexander
  Toshev}, \bibinfo{person}{Samy Bengio}, {and} \bibinfo{person}{Dumitru
  Erhan}.} \bibinfo{year}{2015}\natexlab{}.
\newblock \showarticletitle{Show and tell: A neural image caption generator}.
  In \bibinfo{booktitle}{\emph{Proceedings of the CVPR}}.
  \bibinfo{pages}{3156--3164}.
\newblock


\bibitem[\protect\citeauthoryear{Wang, Yang, Xu, Hanjalic, and Shen}{Wang
  et~al\mbox{.}}{2017}]%
        {wang2017adversarial}
\bibfield{author}{\bibinfo{person}{Bokun Wang}, \bibinfo{person}{Yang Yang},
  \bibinfo{person}{Xing Xu}, \bibinfo{person}{Alan Hanjalic}, {and}
  \bibinfo{person}{Heng~Tao Shen}.} \bibinfo{year}{2017}\natexlab{}.
\newblock \showarticletitle{Adversarial cross-modal retrieval}. In
  \bibinfo{booktitle}{\emph{Proceedings of the 25th ACM MM}}.
  \bibinfo{pages}{154--162}.
\newblock


\bibitem[\protect\citeauthoryear{Wang, He, Wang, Wang, and Tan}{Wang
  et~al\mbox{.}}{2015}]%
        {wang2015}
\bibfield{author}{\bibinfo{person}{Kaiye Wang}, \bibinfo{person}{Ran He},
  \bibinfo{person}{Liang Wang}, \bibinfo{person}{Wei Wang}, {and}
  \bibinfo{person}{Tieniu Tan}.} \bibinfo{year}{2015}\natexlab{}.
\newblock \showarticletitle{Joint feature selection and subspace learning for
  cross-modal retrieval}.
\newblock \bibinfo{journal}{\emph{IEEE transactions on pattern analysis and
  machine intelligence}} \bibinfo{volume}{38}, \bibinfo{number}{10}
  (\bibinfo{year}{2015}), \bibinfo{pages}{2010--2023}.
\newblock


\bibitem[\protect\citeauthoryear{Wang, Yin, Wang, Wu, and Wang}{Wang
  et~al\mbox{.}}{2016b}]%
        {wang2016comprehensive}
\bibfield{author}{\bibinfo{person}{Kaiye Wang}, \bibinfo{person}{Qiyue Yin},
  \bibinfo{person}{Wei Wang}, \bibinfo{person}{Shu Wu}, {and}
  \bibinfo{person}{Liang Wang}.} \bibinfo{year}{2016}\natexlab{b}.
\newblock \showarticletitle{A comprehensive survey on cross-modal retrieval}.
\newblock \bibinfo{journal}{\emph{arXiv preprint arXiv:1607.06215}}
  (\bibinfo{year}{2016}).
\newblock


\bibitem[\protect\citeauthoryear{Wang, Li, and Lazebnik}{Wang
  et~al\mbox{.}}{2016a}]%
        {wang2016learning}
\bibfield{author}{\bibinfo{person}{Liwei Wang}, \bibinfo{person}{Yin Li}, {and}
  \bibinfo{person}{Svetlana Lazebnik}.} \bibinfo{year}{2016}\natexlab{a}.
\newblock \showarticletitle{Learning deep structure-preserving image-text
  embeddings}. In \bibinfo{booktitle}{\emph{Proceedings of the CVPR}}.
  \bibinfo{pages}{5005--5013}.
\newblock


\bibitem[\protect\citeauthoryear{Wang and Livescu}{Wang and Livescu}{2015}]%
        {wang2015large}
\bibfield{author}{\bibinfo{person}{Weiran Wang} {and} \bibinfo{person}{Karen
  Livescu}.} \bibinfo{year}{2015}\natexlab{}.
\newblock \showarticletitle{Large-scale approximate kernel canonical
  correlation analysis}.
\newblock \bibinfo{journal}{\emph{arXiv preprint arXiv:1511.04773}}
  (\bibinfo{year}{2015}).
\newblock


\bibitem[\protect\citeauthoryear{Wang, Liu, Li, Sheng, Yan, Wang, and
  Shao}{Wang et~al\mbox{.}}{2019}]%
        {wang2019camp}
\bibfield{author}{\bibinfo{person}{Zihao Wang}, \bibinfo{person}{Xihui Liu},
  \bibinfo{person}{Hongsheng Li}, \bibinfo{person}{Lu Sheng},
  \bibinfo{person}{Junjie Yan}, \bibinfo{person}{Xiaogang Wang}, {and}
  \bibinfo{person}{Jing Shao}.} \bibinfo{year}{2019}\natexlab{}.
\newblock \showarticletitle{Camp: Cross-modal adaptive message passing for
  text-image retrieval}. In \bibinfo{booktitle}{\emph{Proceedings of the
  ICCV}}. \bibinfo{pages}{5764--5773}.
\newblock


\bibitem[\protect\citeauthoryear{Wu, Jing, Wu, Ji, Dong, Luo, Huang, and
  Wang}{Wu et~al\mbox{.}}{2020}]%
        {wu2020modality}
\bibfield{author}{\bibinfo{person}{Fei Wu}, \bibinfo{person}{Xiao-Yuan Jing},
  \bibinfo{person}{Zhiyong Wu}, \bibinfo{person}{Yimu Ji},
  \bibinfo{person}{Xiwei Dong}, \bibinfo{person}{Xiaokai Luo},
  \bibinfo{person}{Qinghua Huang}, {and} \bibinfo{person}{Ruchuan Wang}.}
  \bibinfo{year}{2020}\natexlab{}.
\newblock \showarticletitle{Modality-specific and shared generative adversarial
  network for cross-modal retrieval}.
\newblock \bibinfo{journal}{\emph{Pattern Recognition}} (\bibinfo{year}{2020}),
  \bibinfo{pages}{107335}.
\newblock


\bibitem[\protect\citeauthoryear{Wu, Lin, and Zha}{Wu et~al\mbox{.}}{2017}]%
        {wu2017joint}
\bibfield{author}{\bibinfo{person}{Jianlong Wu}, \bibinfo{person}{Zhouchen
  Lin}, {and} \bibinfo{person}{Hongbin Zha}.} \bibinfo{year}{2017}\natexlab{}.
\newblock \showarticletitle{Joint latent subspace learning and regression for
  cross-modal retrieval}. In \bibinfo{booktitle}{\emph{Proceedings of the
  SIGIR}}. \bibinfo{pages}{917--920}.
\newblock


\bibitem[\protect\citeauthoryear{Wu, Wang, and Shao}{Wu et~al\mbox{.}}{2018}]%
        {wu2018cycle}
\bibfield{author}{\bibinfo{person}{Lin Wu}, \bibinfo{person}{Yang Wang}, {and}
  \bibinfo{person}{Ling Shao}.} \bibinfo{year}{2018}\natexlab{}.
\newblock \showarticletitle{Cycle-consistent deep generative hashing for
  cross-modal retrieval}.
\newblock \bibinfo{journal}{\emph{IEEE Transactions on Image Processing}}
  \bibinfo{volume}{28}, \bibinfo{number}{4} (\bibinfo{year}{2018}),
  \bibinfo{pages}{1602--1612}.
\newblock


\bibitem[\protect\citeauthoryear{Xia, Pan, Lai, Liu, and Yan}{Xia
  et~al\mbox{.}}{2014}]%
        {xia2014supervised}
\bibfield{author}{\bibinfo{person}{Rongkai Xia}, \bibinfo{person}{Yan Pan},
  \bibinfo{person}{Hanjiang Lai}, \bibinfo{person}{Cong Liu}, {and}
  \bibinfo{person}{Shuicheng Yan}.} \bibinfo{year}{2014}\natexlab{}.
\newblock \showarticletitle{Supervised hashing for image retrieval via image
  representation learning.}. In \bibinfo{booktitle}{\emph{Proceedings of the
  AAAI}}. \bibinfo{pages}{2156--2162}.
\newblock


\bibitem[\protect\citeauthoryear{Xu, Ba, Kiros, Cho, Courville, Salakhudinov,
  Zemel, and Bengio}{Xu et~al\mbox{.}}{2015}]%
        {xu2015show}
\bibfield{author}{\bibinfo{person}{Kelvin Xu}, \bibinfo{person}{Jimmy Ba},
  \bibinfo{person}{Ryan Kiros}, \bibinfo{person}{Kyunghyun Cho},
  \bibinfo{person}{Aaron Courville}, \bibinfo{person}{Ruslan Salakhudinov},
  \bibinfo{person}{Rich Zemel}, {and} \bibinfo{person}{Yoshua Bengio}.}
  \bibinfo{year}{2015}\natexlab{}.
\newblock \showarticletitle{Show, attend and tell: Neural image caption
  generation with visual attention}. In \bibinfo{booktitle}{\emph{Proceedings
  of the ICML}}. \bibinfo{pages}{2048--2057}.
\newblock


\bibitem[\protect\citeauthoryear{Yang, Zhang, Yin, and Liu}{Yang
  et~al\mbox{.}}{2018}]%
        {yang2018robust}
\bibfield{author}{\bibinfo{person}{Hong-Ming Yang}, \bibinfo{person}{Xu-Yao
  Zhang}, \bibinfo{person}{Fei Yin}, {and} \bibinfo{person}{Cheng-Lin Liu}.}
  \bibinfo{year}{2018}\natexlab{}.
\newblock \showarticletitle{Robust classification with convolutional prototype
  learning}. In \bibinfo{booktitle}{\emph{Proceedings of the CVPR}}.
  \bibinfo{pages}{3474--3482}.
\newblock


\bibitem[\protect\citeauthoryear{Zeng, Sun, and Mao}{Zeng
  et~al\mbox{.}}{2021a}]%
        {zeng2021mccn}
\bibfield{author}{\bibinfo{person}{Zhixiong Zeng}, \bibinfo{person}{Ying Sun},
  {and} \bibinfo{person}{Wenji Mao}.} \bibinfo{year}{2021}\natexlab{a}.
\newblock \showarticletitle{MCCN: Multimodal Coordinated Clustering Network for
  Large-Scale Cross-modal Retrieval}. In \bibinfo{booktitle}{\emph{Proceedings
  of the 29th ACM International Conference on Multimedia}}.
  \bibinfo{pages}{5427--5435}.
\newblock


\bibitem[\protect\citeauthoryear{Zeng, Wang, Xu, and Mao}{Zeng
  et~al\mbox{.}}{2021b}]%
        {zeng2021pan}
\bibfield{author}{\bibinfo{person}{Zhixiong Zeng}, \bibinfo{person}{Shuai
  Wang}, \bibinfo{person}{Nan Xu}, {and} \bibinfo{person}{Wenji Mao}.}
  \bibinfo{year}{2021}\natexlab{b}.
\newblock \showarticletitle{PAN: Prototype-based Adaptive Network for Robust
  Cross-modal Retrieval}. In \bibinfo{booktitle}{\emph{Proceedings of the
  SIGIR}}. \bibinfo{pages}{1125--1134}.
\newblock


\bibitem[\protect\citeauthoryear{Zeng, Xu, and Mao}{Zeng et~al\mbox{.}}{2020}]%
        {zeng2020event}
\bibfield{author}{\bibinfo{person}{Zhixiong Zeng}, \bibinfo{person}{Nan Xu},
  {and} \bibinfo{person}{Wenji Mao}.} \bibinfo{year}{2020}\natexlab{}.
\newblock \showarticletitle{Event-Driven Network for Cross-Modal Retrieval}. In
  \bibinfo{booktitle}{\emph{Proceedings of the 29th ACM International
  Conference on Information \& Knowledge Management}}.
  \bibinfo{pages}{2297--2300}.
\newblock


\bibitem[\protect\citeauthoryear{Zhai, Peng, and Xiao}{Zhai
  et~al\mbox{.}}{2013}]%
        {zhai2013learning}
\bibfield{author}{\bibinfo{person}{Xiaohua Zhai}, \bibinfo{person}{Yuxin Peng},
  {and} \bibinfo{person}{Jianguo Xiao}.} \bibinfo{year}{2013}\natexlab{}.
\newblock \showarticletitle{Learning cross-media joint representation with
  sparse and semisupervised regularization}.
\newblock \bibinfo{journal}{\emph{IEEE Transactions on Circuits and Systems for
  Video Technology}} \bibinfo{volume}{24}, \bibinfo{number}{6}
  (\bibinfo{year}{2013}), \bibinfo{pages}{965--978}.
\newblock


\bibitem[\protect\citeauthoryear{Zhang, Lei, Zhang, and Li}{Zhang
  et~al\mbox{.}}{2020}]%
        {zhang2020context}
\bibfield{author}{\bibinfo{person}{Qi Zhang}, \bibinfo{person}{Zhen Lei},
  \bibinfo{person}{Zhaoxiang Zhang}, {and} \bibinfo{person}{Stan~Z Li}.}
  \bibinfo{year}{2020}\natexlab{}.
\newblock \showarticletitle{Context-aware attention network for image-text
  retrieval}. In \bibinfo{booktitle}{\emph{Proceedings of the CVPR}}.
  \bibinfo{pages}{3536--3545}.
\newblock


\bibitem[\protect\citeauthoryear{Zhang}{Zhang}{2019}]%
        {zhang2019making}
\bibfield{author}{\bibinfo{person}{Richard Zhang}.}
  \bibinfo{year}{2019}\natexlab{}.
\newblock \showarticletitle{Making convolutional networks shift-invariant
  again}. In \bibinfo{booktitle}{\emph{International conference on machine
  learning}}. PMLR, \bibinfo{pages}{7324--7334}.
\newblock


\bibitem[\protect\citeauthoryear{Zhen, Hu, Wang, and Peng}{Zhen
  et~al\mbox{.}}{2019}]%
        {zhen2019deep}
\bibfield{author}{\bibinfo{person}{Liangli Zhen}, \bibinfo{person}{Peng Hu},
  \bibinfo{person}{Xu Wang}, {and} \bibinfo{person}{Dezhong Peng}.}
  \bibinfo{year}{2019}\natexlab{}.
\newblock \showarticletitle{Deep Supervised Cross-Modal Retrieval}. In
  \bibinfo{booktitle}{\emph{Proceedings of the CVPR}}.
  \bibinfo{pages}{10394--10403}.
\newblock


\bibitem[\protect\citeauthoryear{Zheng, Tang, and Shao}{Zheng
  et~al\mbox{.}}{2016}]%
        {zheng2016hetero}
\bibfield{author}{\bibinfo{person}{Feng Zheng}, \bibinfo{person}{Yi Tang},
  {and} \bibinfo{person}{Ling Shao}.} \bibinfo{year}{2016}\natexlab{}.
\newblock \showarticletitle{Hetero-manifold regularisation for cross-modal
  hashing}.
\newblock \bibinfo{journal}{\emph{IEEE transactions on pattern analysis and
  machine intelligence}} \bibinfo{volume}{40}, \bibinfo{number}{5}
  (\bibinfo{year}{2016}), \bibinfo{pages}{1059--1071}.
\newblock


\end{thebibliography}

\end{document}